\documentclass[runningheads]{llncs}

\usepackage[width=122mm,left=12mm,paperwidth=146mm,height=193mm,top=12mm,paperheight=217mm]{geometry}
\usepackage[numbers]{natbib}
\usepackage{multicol}
\usepackage[bookmarks=true]{hyperref}
\usepackage{times}
\usepackage{epsfig}
\usepackage{graphicx}
\usepackage{amsmath}
\usepackage{amssymb}
\usepackage{multirow}
\usepackage{booktabs}
\usepackage{bbm}
\usepackage[toc,page]{appendix}

\usepackage{colortbl}
\usepackage[usenames, dvipsnames]{xcolor}
\usepackage{xspace}
\usepackage{ifpdf}
\usepackage{xstring}

\newcommand{\UP}[1]{(\IfSubStr{#1}{-}{ \textcolor{red}{\bf #1} }{ \textcolor{blue}{\bf #1}})}

\def\eg{\emph{e.g.~}}
\def\etal{\emph{et al.~}}

\def\ie{\emph{i.e.~}}

\begin{document}
% \renewcommand\thelinenumber{\color[rgb]{0.2,0.5,0.8}\normalfont\sffamily\scriptsize\arabic{linenumber}\color[rgb]{0,0,0}}
% \renewcommand\makeLineNumber {\hss\thelinenumber\ \hspace{6mm} \rlap{\hskip\textwidth\ \hspace{6.5mm}\thelinenumber}}
% \linenumbers
\pagestyle{headings}
\mainmatter
\def\ECCV16SubNumber{1305}  % Insert your submission number here
	
%\title{Knowledge Transfer and Model Compression for Deconvolutional Networks with\\ Multi-Domain Training}
%\title{Multi-Domain Training and Knowledge Distillation for Efficient Deconvolutional Networks in Street-View Semantic Segmentation}
%\title{Recipes for Training Deconvolutional Networks for Semantic Segmentation in Automotive Environments}
\title{Training Constrained Deconvolutional Networks for Road Scene Semantic Segmentation}

\titlerunning{submitted as a conference paper}
\authorrunning{submitted as a conference paper}
\author{German Ros\inst{1}\thanks{ This work was done as part of first author’s internships at Toshiba Research Europe and Toshiba Corporate Research \& Development Center. Pablo F. Alcantarilla contributions to this work were done while he was working at Toshiba Research Europe. We thank Toshiba for its support during the realization of this project.}
 \and Simon Stent\inst{2} \and Pablo F. Alcantarilla\inst{3} \and Tomoki Watanabe\inst{4}}

\institute{
Computer Vision Center, UAB, Barcelona, Spain\\
\email{gros@cvc.uab.es}\\
\and
Department of Engineering, Cambridge University, Cambridge, UK\\
\email{sais2@eng.cam.ac.uk}\\
\and
iRobot Corporation, London, UK\\
\email{palcantarilla@irobot.com}\\
\and
 Corporate Research \& Development Center, Toshiba Corporation, Kawasaki, Japan\\
\email{tomoki8.watanabe@toshiba.co.jp}\\
}

%\titlerunning{Training Practical Deconvolutional Networks for Semantic Road Scene Segmentation}
%\authorrunning{}
%\author{German Ros \and Simon Stent \and Pablo Fernandez \and Tomoki Watanabe}
%\institute{Toshiba Research and Development Corporation}

\maketitle

\begin{abstract}
In this work we investigate the problem of road scene semantic segmentation using Deconvolutional Networks (DNs). Several constraints limit the practical performance of DNs in this context: firstly, the paucity of existing pixel-wise labelled training data, and secondly, the memory constraints of embedded hardware,  which rule out the practical use of state-of-the-art DN architectures such as fully convolutional networks (FCN). 
To address the first constraint, we introduce a Multi-Domain Road Scene Semantic Segmentation (MDRS3) dataset, aggregating data from six existing densely and sparsely labelled datasets for training our models, and two existing, separate datasets for testing their generalisation performance. We show that, while MDRS3 offers a greater volume and variety of data, end-to-end training of a memory efficient DN does not yield satisfactory performance. We propose a new training strategy to overcome this, based on (i) the creation of a best-possible source network (S-Net) from the aggregated data, ignoring time and memory constraints; and (ii) the transfer of knowledge from S-Net to the memory-efficient target network (T-Net).
We evaluate different techniques for S-Net creation and T-Net transferral, and demonstrate that training a constrained deconvolutional network in this manner can unlock better performance than existing training approaches. Specifically, we show that a target network can be trained to achieve improved accuracy versus an FCN despite using less than 1\% of the memory. We believe that our approach can be useful beyond automotive scenarios where labelled data is similarly scarce or fragmented and where practical constraints exist on the desired model size. We make available our network models and aggregated multi-domain dataset for reproducibility.

\end{abstract}
	
%*******************************************************************************
\section{Introduction}\label{sec:intro}
\vspace{-2mm}
Deconvolutional Networks (DNs) are a class of neural network which have achieved notable recent success on the task of semantic segmentation, in which image recognition is performed at the resolution of individual pixels~\cite{Farabet13pami,Long15cvpr,Noh15iccv,Badrinarayanan15arxiv}. They have consequently become an attractive architecture for road scene segmentation---a useful component in many autonomous driving or advanced driver assistance systems. However, several limitations exist when trying to apply state-of-the-art DNs in practice.

Firstly, they are inefficient in terms of memory footprint. While commercial chips targeting the automotive industry are becoming increasingly parallel, the small size of fast-access on-chip SRAM memories remains limited (\eg 512 KB for the Mobileye EyeQ2\footnote{\url{http://www.mobileye.com/technology/processing-platforms/eyeq2/}}\footnote{\label{footnote:disclaimer}All product names may be trademarks of their respective companies.} chip and 1-10 MB for the Toshiba TMPV 760 Series\footnote{\url{http://toshiba.semicon-storage.com/ap-en/product/automotive.html}}\textsuperscript{\ref{footnote:disclaimer}} chip family). In contrast, the popular FCN-8s network~\cite{Long15cvpr} with 134.5 M parameters requires more than 500 MB of memory. %been benchmarked with a frame-rate of 4.8 Hz for resolutions of $500{\times}500$ pixels and . 
Although more efficient architectures have been proposed, such as SegNet~\cite{Badrinarayanan15arxiv}, they still contain tens of millions of parameters (29.5 M for~\cite{Badrinarayanan15arxiv}) and are yet to demonstrate accuracy on a par with the larger FCN-8s.
%--benchmarked at 10.6 Hz and 

Secondly, since DNs are typically trained in a supervised manner, their performance benefits from access to a large amount of training data with corresponding per-pixel annotations. Producing such annotations is an expensive and time-consuming process. Hence, while datasets for tasks such as image classification can reach $\mathcal{O}(10^7)$ images in scale~\cite{Lin14eccv,Russakovsky15ijcv,MITPLACES}, popular semantic road scene segmentation datasets such as CamVid~\cite{Brostow09prl} or KITTI~\cite{Geiger13ijrr} contain $\mathcal{O}(10^3)$ images. The scarcity of data results in a lack of samples for rarer but important classes such as pedestrians and cyclists, which can make it difficult for models to learn these concepts without overfitting. Furthermore, data scarcity implies poor coverage over the true distribution of possible road scenes: datasets are typically captured in one or a few localised regions under relatively homogeneous road conditions. Understanding how best to incorporate knowledge from new domains as training data becomes available is an important problem to ensure the best general task performance given available data.

\begin{figure}[t!]
	\centering
	\includegraphics[width=1.\textwidth]{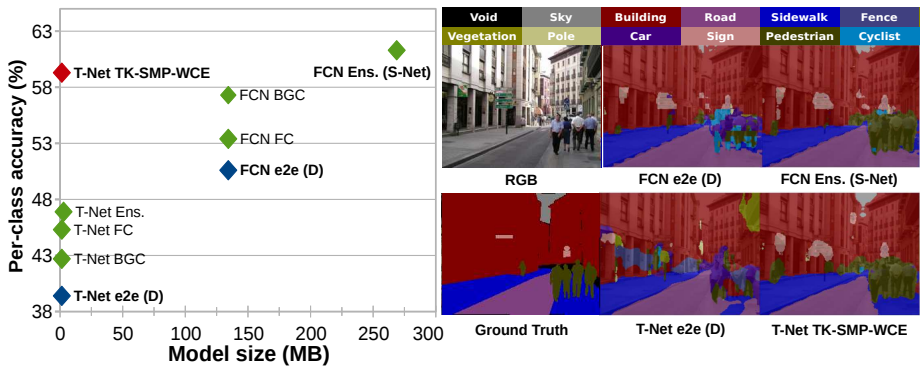}
	\vspace{-7mm}
	\caption{\textit{Left:} Model performance versus model size. Blue points indicate the baseline performance of the two different network architectures we compare. Green points indicate performance using different training strategies explained in Sec.~\ref{sec:training}. Red point denotes the performance of our constrained network after knowledge transfer, explained in Sec.~\ref{sec:training}. Our approach outperforms the per-class accuracy of a state-of-the-art fully convolutional network (FCN e2e (D)) with 1\% of the model size. \textit{Right:} Segmentation output on a test image, illustrating the qualitative improvement of our method. Best viewed on screen.}
	%\textbf{\textcolor{Blue}{Blue}} data points refer to simple end-to-end training on the dense-only training sets of our two baseline architectures: T-Net, our memory efficient production network architecture, and FCN, the fully convolutional network of~\cite{Long15cvpr}. %Training end-to-end on combined dense and sparse datasets resulted in poor performance as explained in Section \ref{sec:exps}.
	%\textbf{\textcolor{ForestGreen}{Green}} points show the performance of training strategies explored to boost the performance of the baseline networks, as explained in Section~\ref{sec:training}.
	%\textbf{\textcolor{BrickRed}{Red}} points show the performance of our final production networks, trained via knowledge transfer as described in Section~\ref{sec:transfer}. Our proposed training strategy allows better control than existing training methods over the final trade-off between task performance, model size and computation time.}
	\label{fig:summary}
		\vspace{-4mm}
\end{figure}

To address these limitations, we propose an approach which draws on ideas from domain adaptation~\cite{Chen15cvpr,Tzeng15iccv} and model 
compression by transfer learning~\cite{HintonVinyals15}. We begin by collating numerous publicly available datasets from different domains 
and modalities which are useful for the task of semantic road scene segmentation. We refer to our aggregated dataset as the Multi-Domain 
Road Scene Semantic Segmentation (MDRS3) dataset. In contrast to existing work in road scene semantic 
segmentation~\cite{Kundu14eccv,Ros15wacv}, we select two of the constituent datasets in their entirety as the test set for MDRS3. This 
means that training and testing for MDRS3 are not carried out on subsets of the same original dataset and performance is a better indication 
of task generalisation. We then examine methods for training on different modalities of data to create the best possible model, ignoring 
time and memory constraints. We discover that ensembling networks trained on distinct domains leads to much improved performance, and create 
a best-performing network containing 269 million parameters, which we refer to as the Source Network (S-Net). Finally, we explore 
methods for transferring knowledge from the unconstrained S-Net to a memory constrained architecture, which we refer to as the 
Target Network (T-Net). We demonstrate that by this approach, we can meet the desired constraints for embedded applications while achieving 
a higher accuracy than is otherwise impossible through existing training strategies. Concretely, we show that the performance of a 
state-of-the-art FCN~\cite{Long15cvpr} can be bettered using a deconvolutional network with 1\% of the memory and comparable run-time. 
Fig.~\ref{fig:summary} summarises our experimental findings. 
For reproducibility, we plan to make publicly available all of our trained models and MDRS3 dataset upon publication.% with several hundreds of new pixelwise 
%annotations alongside this paper, marked in Fig.~\ref{fig:datasets} with an asterisk. 

\section{Related Work}
\label{sec:related}
\vspace{-1mm}

Our work spans the topics of semantic segmentation, training with limited data, and knowledge transfer. We briefly recapitulate related literature from each.
%\paragraph{\bf{\textnormal{2.1 }}\textbf{On Semantic Segmentation and Data limitations}}\ \vspace{2mm}\newline

\paragraph{\bf Semantic Segmentation.} 
%Formally, semantic segmentation can be described as a mapping from an image $\mathcal{I} \in \mathbb{R}^{H \times W}$ to a predefined set of labels $\mathcal{L}$ (\eg road, sidewalk, sky, vegetation, pedestrians, etc.). This task can be carried out by a DeconvNet $\mathcal{D}:  \mathcal{I} \in \mathbb{R}^{H \times W} \times \theta \rightarrow \mathcal{L}^{H \times W}$ such that $\mathcal{D}(\mathcal{I}, \theta) = \{l_{i,j}\}_{i=1,j=1}^{H,W}$, where $\theta$ is the model or set of trainable parameters.
%\GR{I think adding the formal definition is ok :(}
% ok - added a much simpler definition thouogh... 
The task of semantic segmentation involves the estimation of a function $\mathcal{F}$ which maps an input image, such as $\mathcal{I} \in 
[0,\dots,255]^{H \times W}$, to an output label image $\mathcal{J} \in [1,\dots,L]^{H \times W}$, where the labels $1,\dots,L$ 
index the semantic class of the input at that pixel (\eg road, sidewalk, sky, vegetation, pedestrians, etc.). 
It is a popular problem in computer vision and has been tackled for various environments from 
indoors~\cite{Silberman12,Handa16cvpr} to outdoors~\cite{Kundu14eccv,Tighe10eccv}, as well as for specific tasks such as road 
scene perception~\cite{Sengupta13icra}. For the latter, which is the focus of our work, semantic segmentation is expected to play 
a key role as part of the local planning and obstacle avoidance sub-systems of future semi-autonomous and autonomous vehicles. 

Classical tools for addressing the problem include pipelines based on a combination of hand-crafted features (\eg SIFT, HOG) and 
region-based classifiers (\eg SVM, ADABoost), with probabilistic graphical models such as Conditional Random Fields~(CRFs) used to 
produce structured predictions~\cite{Ros15wacv,Sengupta13icra,Hu13icra,Kohli09ijcv,Ladicky10eccv,Valentin13}. With the arrival of 
deep convolutional neural networks (CNNs), hand-crafted features were substituted  by learned CNN representations, which worked at 
the level of image patches~\cite{Alvarez12eccv,Girshick14rcnn}. This trend continued with the introduction of DNs, which naturally 
perform the process of recognition and whole-image segmentation, producing a dense inference at a pixel 
level~\cite{Long15cvpr,Noh15iccv,Badrinarayanan15arxiv,Silberman12}. Recently, this trend has culminated in the addition of 
structured prediction by employing messaging passing between the net and an MRF~\cite{icml2015_chenb15} or adopting the use of 
Recurrent Nets with equivalent behaviours to a CRF~\cite{Zheng15iccv}. 

%DNs typically consist of a compression network, in which input information is transformed first by compressing into a bottleneck using a traditional convolutional network; followed by an expansion network, in which the deconvolution operation is used to restore the spatial resolution of the compressed representation. Deconvolution during expansion is commonly performed using a simple upsampling approach as in~\cite{Long15cvpr}. in [FCN] performed on coarse activation maps, our algorithm generates object segmentation masks using deep deconvolution network, where a dense pixel-wise class probability map is obtained by successive
%operations of unpooling, deconvolution, and rectification

\paragraph{\bf Training with Limited Data.}
One key problem with DNs is that when applied to certain domains, such as automotive environments, there is a lack of suitably large and varied training data. 
Two recent approaches~\cite{Dai15iccv,Papandreu15iccv} propose means of mitigating this problem by augmenting an existing semantic segmentation dataset (\ie consisting of pixel-wise labels) with additional data from object detection and image classification datasets, which are weakly annotated with bounding boxes or text captions. Both approaches are directly applied on the augmented datasets to train DNs in an end-to-end fashion and both report subsequent improvements in accuracy for the PASCAL-VOC dataset~\cite{Everingham15}.
However, obtaining significant improvements in this manner is possible only when the existing and additional datasets are similar in nature---in this case, both consisted of annotations of simple objects, ignoring the architectonic elements composing the background, \eg road, sidewalk, etc. As we show in this work, the application of these strategies when dealing with urban imagery and architectonic classes fails to produce the most competitive results in automotive scenarios. 
Furthermore, while the issue of training data scarcity is likely to diminish with time, as new larger datasets are released (\eg the recent releases of the Cityscapes~\cite{Cordts2015Cvprw} dataset, containing 5,000 fine-labelled images and 20,000 coarse-labelled images, and the SYNTHIA dataset~\cite{Gros2016}, containing 250,000 synthetic fine-labelled images),  we believe that our approach will remain useful for training resource-constrained segmentation networks.
% for as long as there is heterogeneity present among \SI{or even within?} datasets (\eg fine-labels, coarse-labels, bounding boxes, synthetic data, different regions, different weather conditions etc.).
%We show how this problem can be avoided by carrying out the training process in several steps. 
% and to the infeasibility of applying a simple ``mix-data and train" approach to such a large collection of data, the ideas exposed in this work can still help to boost model accuracy.

%\SI{no mention of domain adapation anywhere?} \GR{Well, the only DA technique is the flying-cars and perhaps the balanced training...}

%\paragraph{\bf{\textnormal{2.2 }}\textbf{Transfer Knowledge and the limitations of End-to-End training}}\ \vspace{2mm}\newline

\vspace{-2mm}
\paragraph{\bf Model Compression by Knowledge Transfer.}
While the recent trend in deep learning has been to strive for even deeper models~\cite{He15cvpr}, the preference for deep versus shallower models is not because shallower models have been shown to have limited capacity or representational power, but rather that learning and regularization procedures used to train shallow models are not sufficiently powerful~\cite{Ba14nips}. 
%In~\cite{Choromanska15aistats}, the authors show that for small networks, getting stuck in poor local minima during training is a major problem, but that this problem becomes less detrimental as the size of the network increases, since critical points of large networks exhibit a layered structure where high-quality low-index critical points lie close to the global minimum. 
One reason for this is that, counterintuitively, the likelihood of falling into poor quality local minima increases with decreasing network size~\cite{Choromanska15aistats}. 
Various approaches to extract better performance from shallow networks have been proposed in the literature. 
%In~\cite{Bucilua06kdd}, an ensemble of classifiers trained on a small dataset is used to label a large dataset, ideally drawn from the true distribution of interest. 
In~\cite{Bucilua06kdd}, an ensemble of classifiers, trained on a small but representative subset of a larger dataset, is used to label the larger unlabelled dataset. The large ensemble-labelled dataset is then used to train a network, demonstrating improved performance versus training on the original ground truth for the smaller dataset. 
More recently, \citep{Ba14nips} shows that shallow neural nets can be trained to achieve performances previously only reachable by deep models, by training a large teacher ensemble and transferring knowledge from it to a shallow but wide model by training it to match the logit activations of the teacher.
Hinton \etal~\cite{HintonVinyals15} confirm these findings and propose to address the problem by exploiting the ``dark knowledge" available in the teacher ensemble, referring to the full probability distribution produced by the soft-max classifier. This knowledge is transferred to a compact student network using relaxation of cross-entropy. This approach was extended in~\cite{Romero15-iclr}, showing that it is possible to reduce the number of parameters by creating deeper-and-thinner students out of shallower-wider teachers, at the possible expense of increasing computation time. A further recent line of relevant work on network compression focuses on applying sophisticated engineering tools to reduce the network size. Examples include \cite{HanMD15}, which uses  pruning, trained quantization and Huffman coding to further compression, and ~\cite{Iandola16arxiv}, in which these engineering tools are combined with a novel architecture to produce very compact classification networks.
In this paper we extend previous research on knowledge transfer to a novel problem and the recently proposed architecture of DNs. % by analysing different techniques and proposing a training approach based on balanced-batches and knowledge transfer between very different types of architectures in order to meet production constraints. 

\section{Multi-Domain Road Scene Semantic Segmentation (MDRS3)}\label{sec:datasets}

\begin{figure*}[t]
	\centering
	\resizebox{\textwidth}{!}{%
		\scriptsize
		\tabcolsep=0.042cm
		\begin{tabular}{c c c c c}
		\toprule
			\textbf{DENSE:} & CamVid (2007)~\cite{Brostow09prl,Brostow08eccv} & KITTI-S (2012)~\cite{Geiger13ijrr} & *Urban LabelMe (2008)~\cite{Russell08ijcv} & 			*CBCL (2013)~\cite{Bileschi07cbcl} \\ \midrule
			\textit{Example(s)} &
				\includegraphics[width=0.09\textwidth]{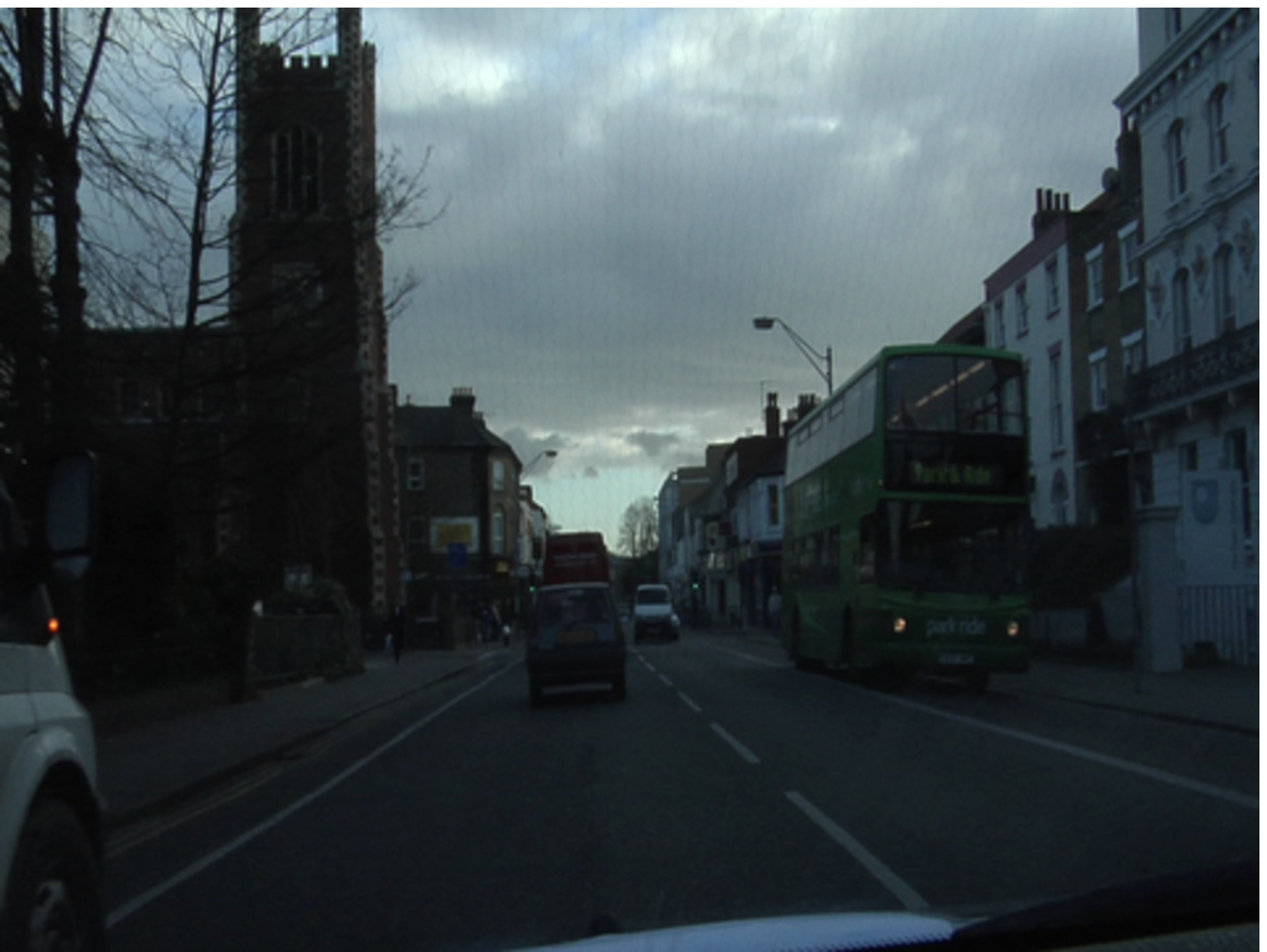}
				\includegraphics[width=0.09\textwidth]{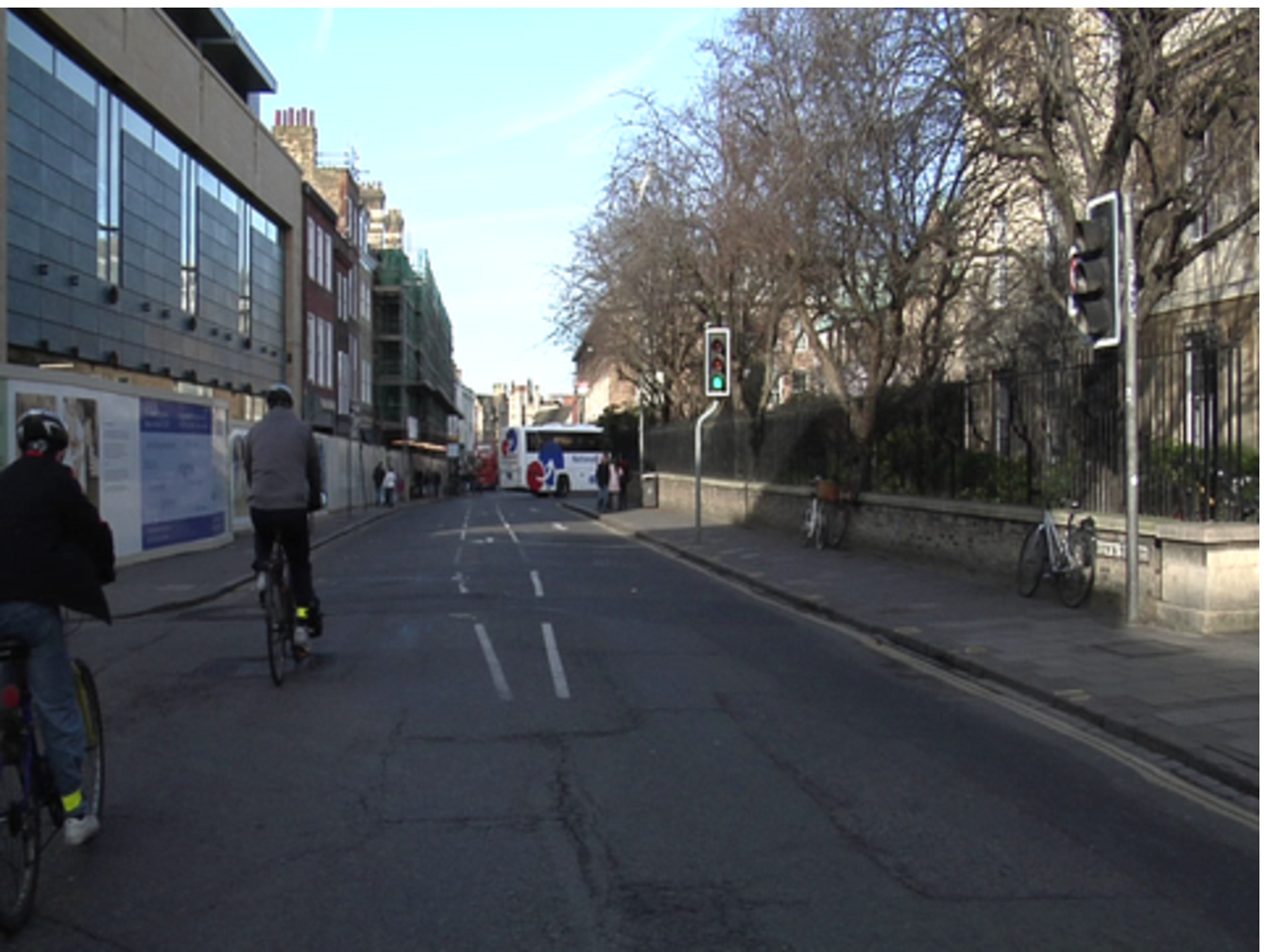} & 
				\includegraphics[width=0.18\textwidth]{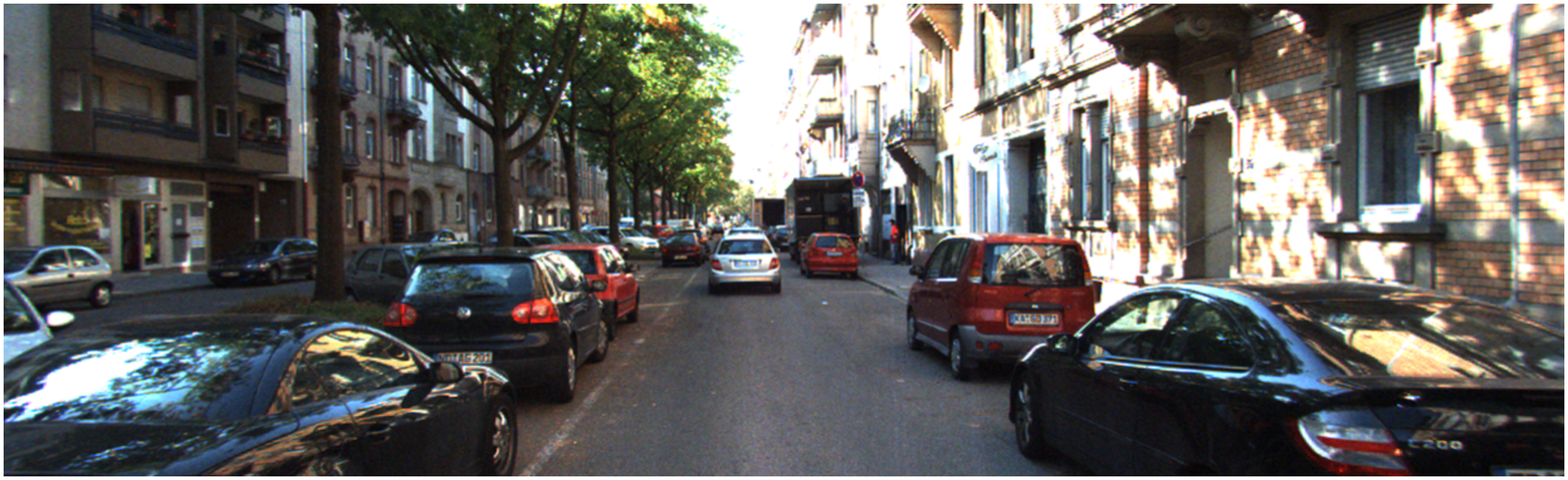} &
				\includegraphics[width=0.10\textwidth]{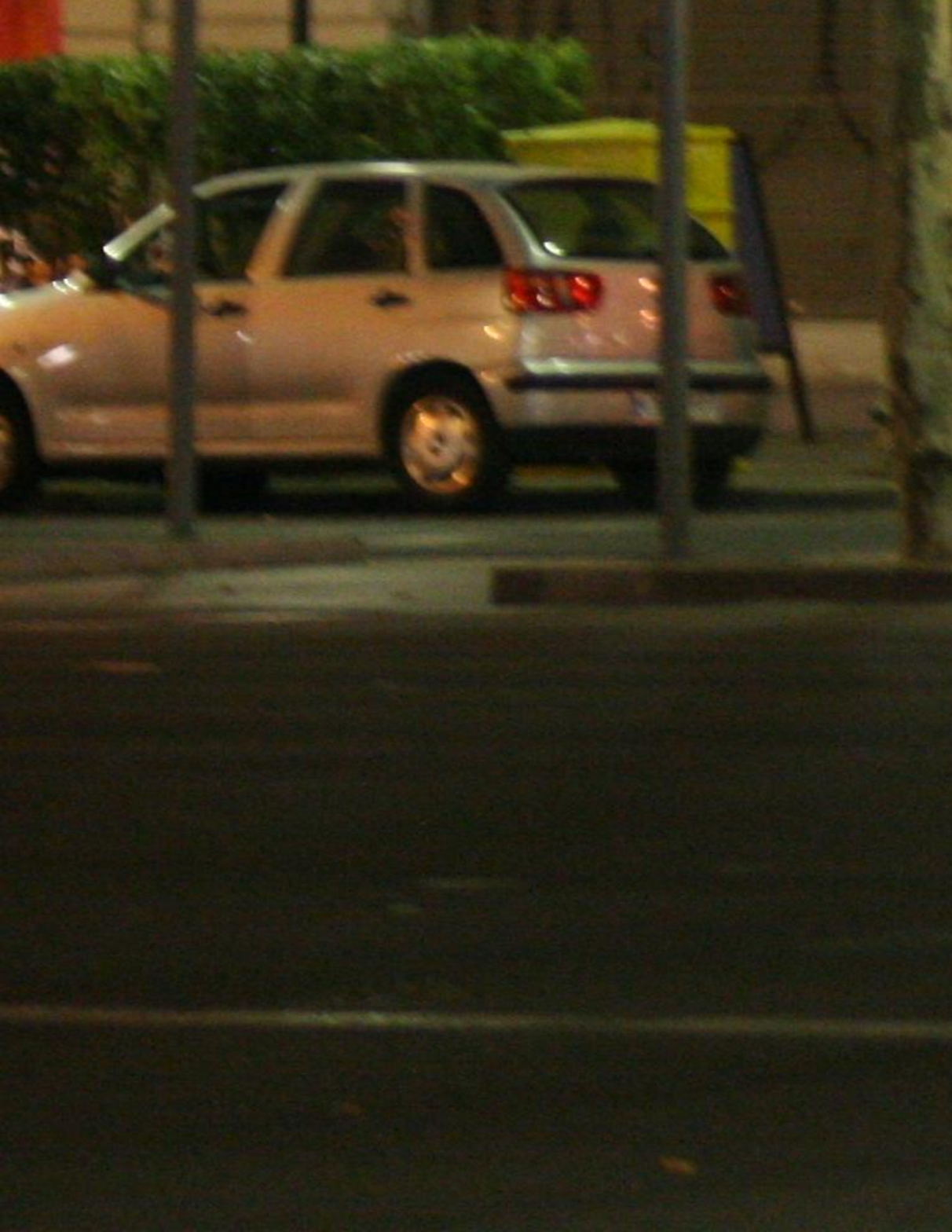}
				\includegraphics[width=0.10\textwidth]{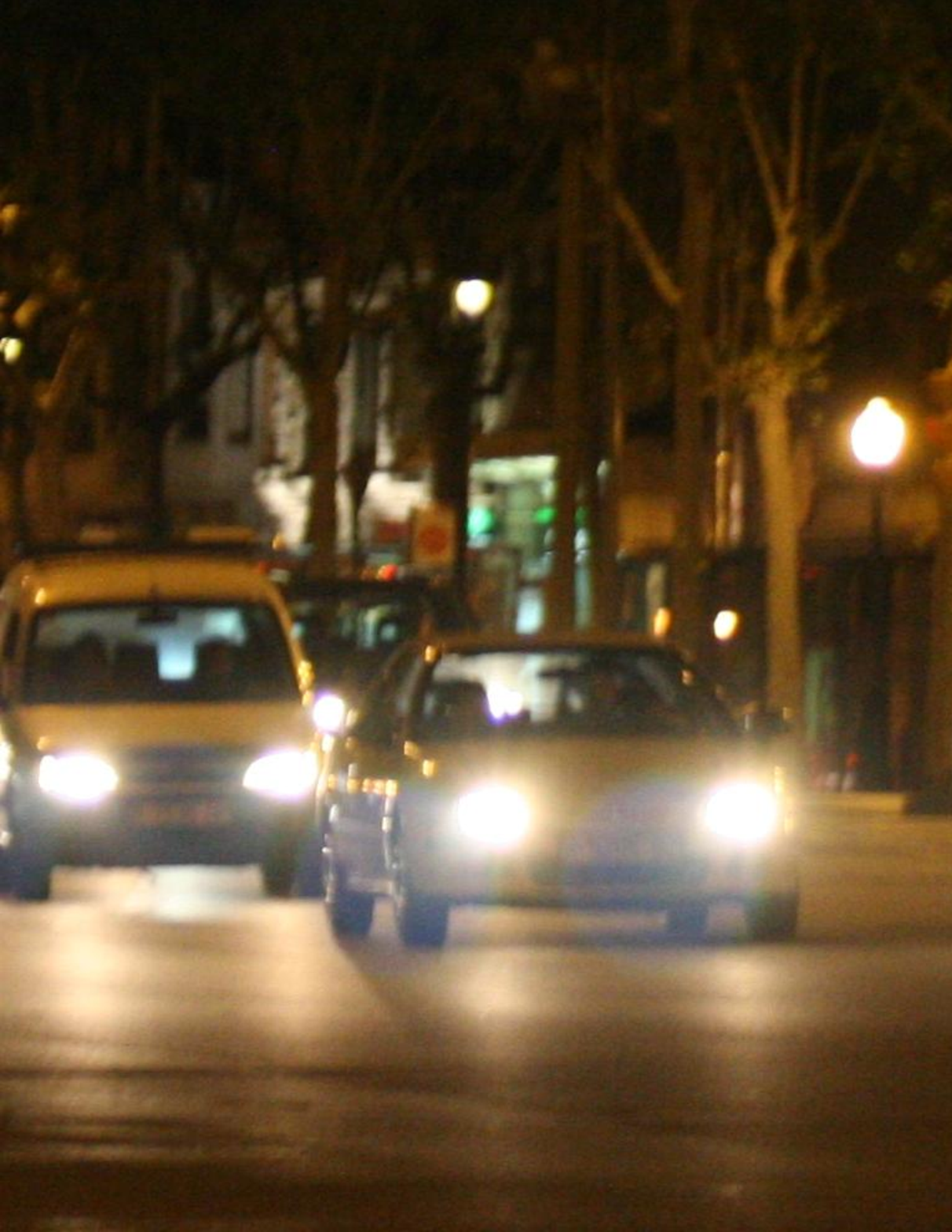} &
				\includegraphics[width=0.09\textwidth]{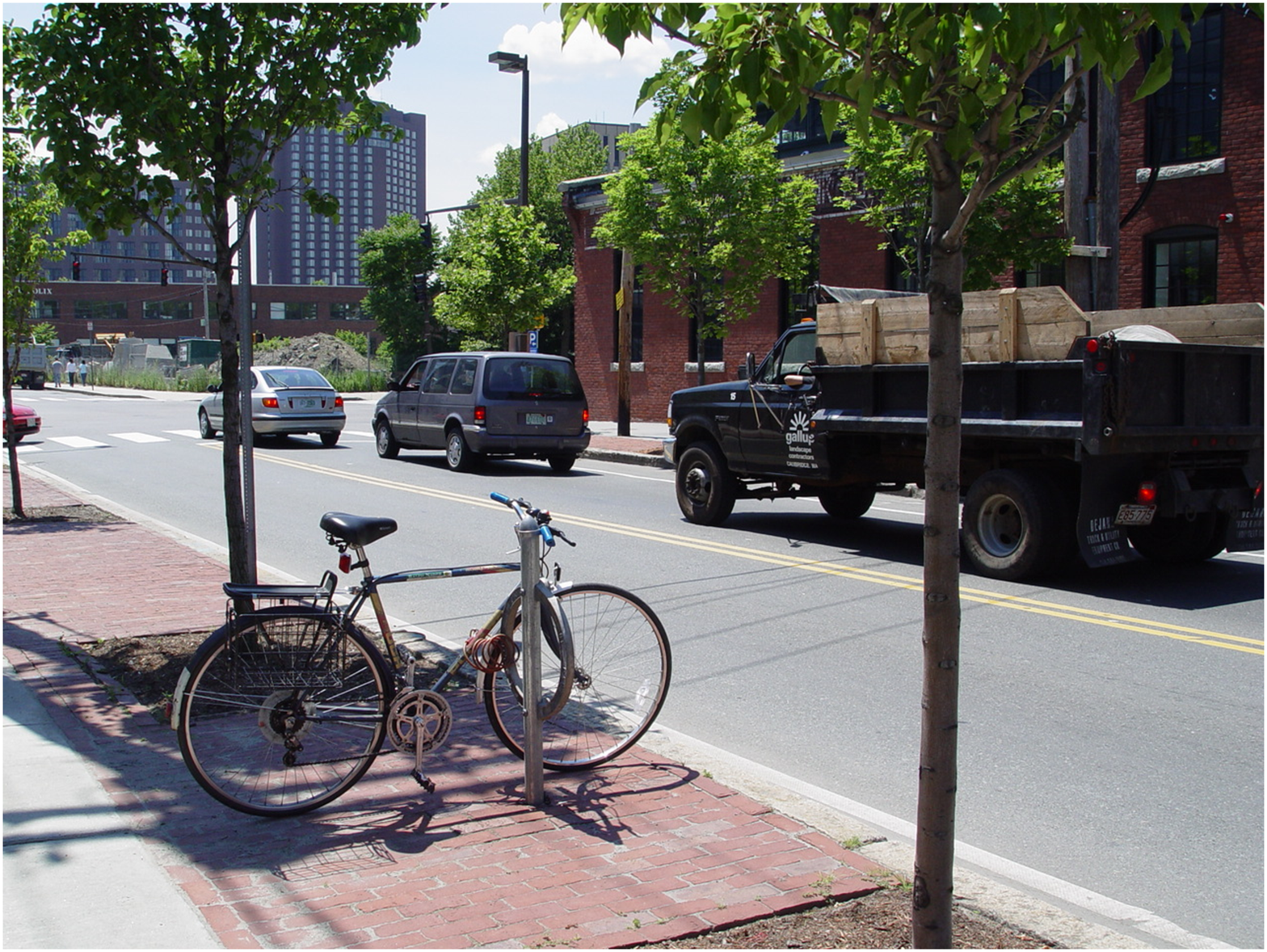}
				\includegraphics[width=0.09\textwidth]{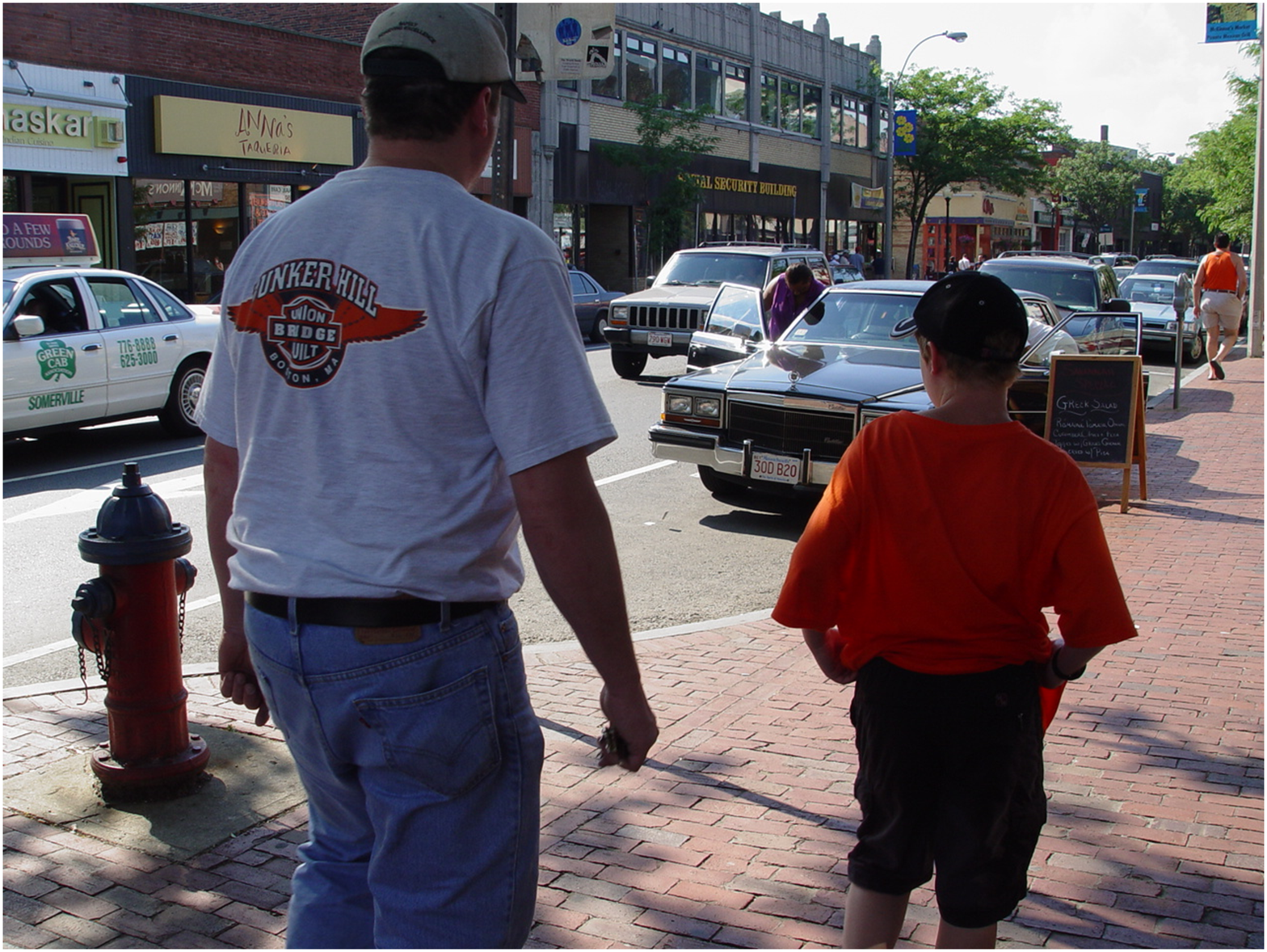} \\
			%\textit{Ground truth} &
			&
				\includegraphics[width=0.09\textwidth]{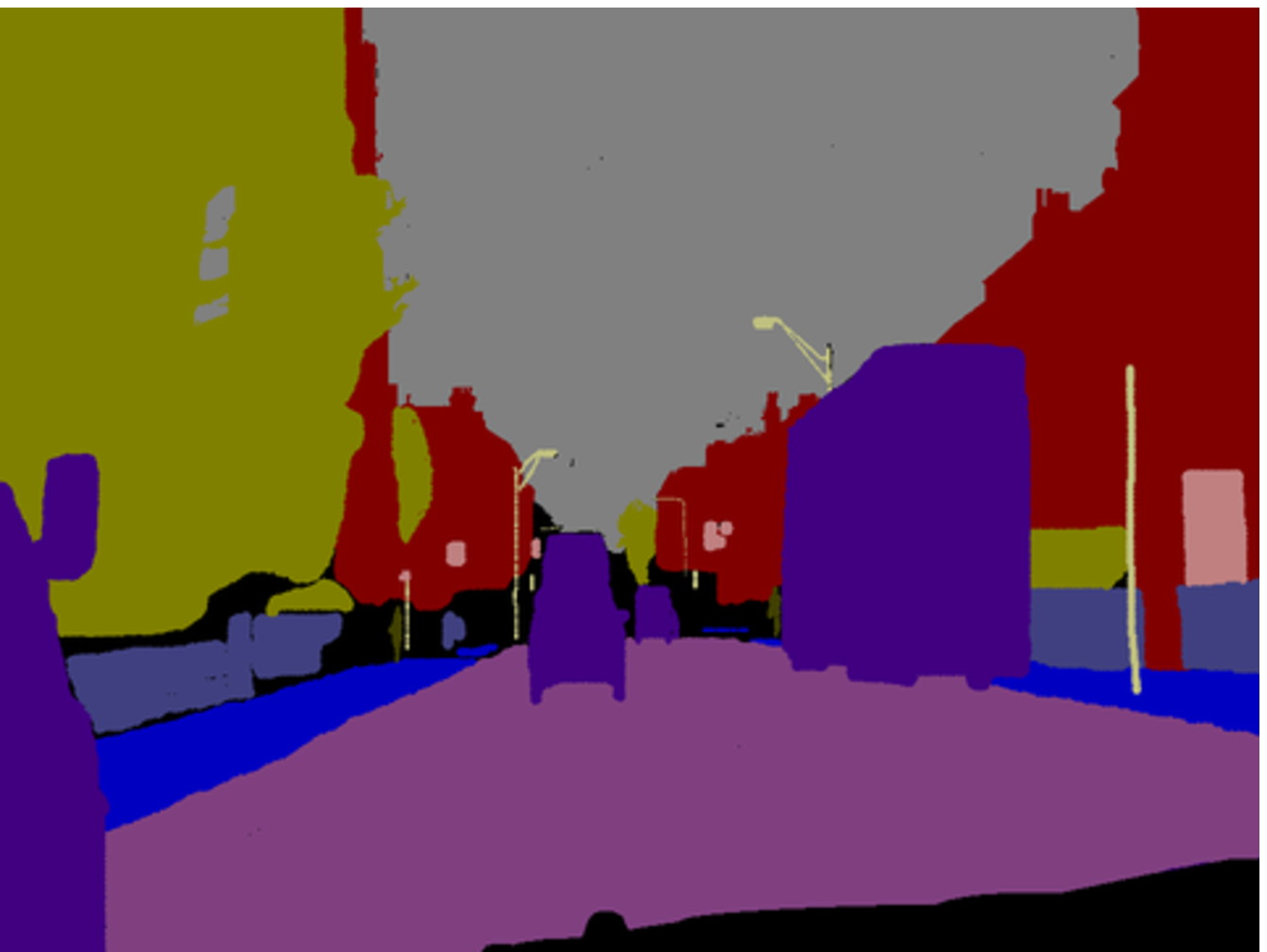}
				\includegraphics[width=0.09\textwidth]{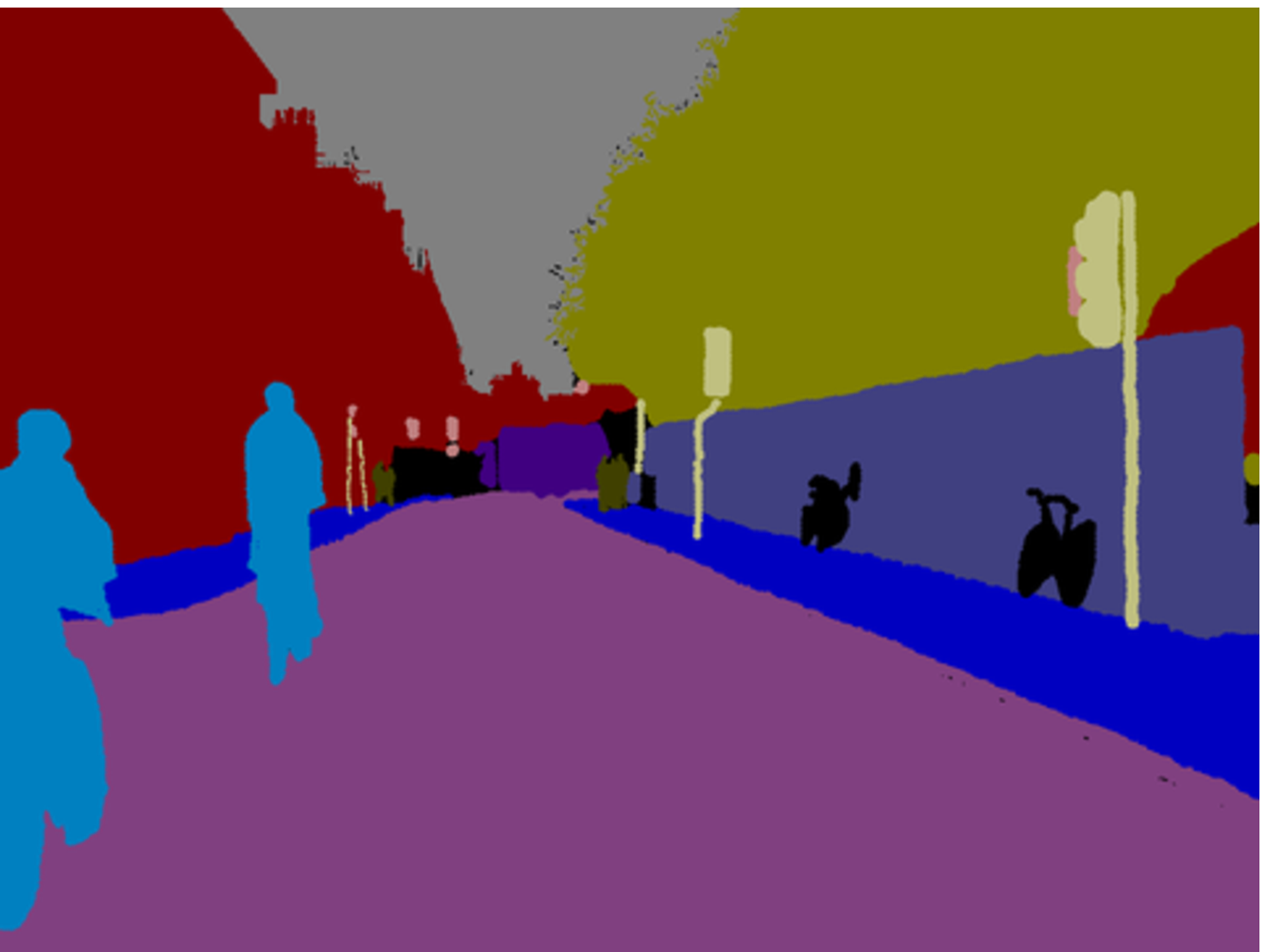} & 
				\includegraphics[width=0.18\textwidth]{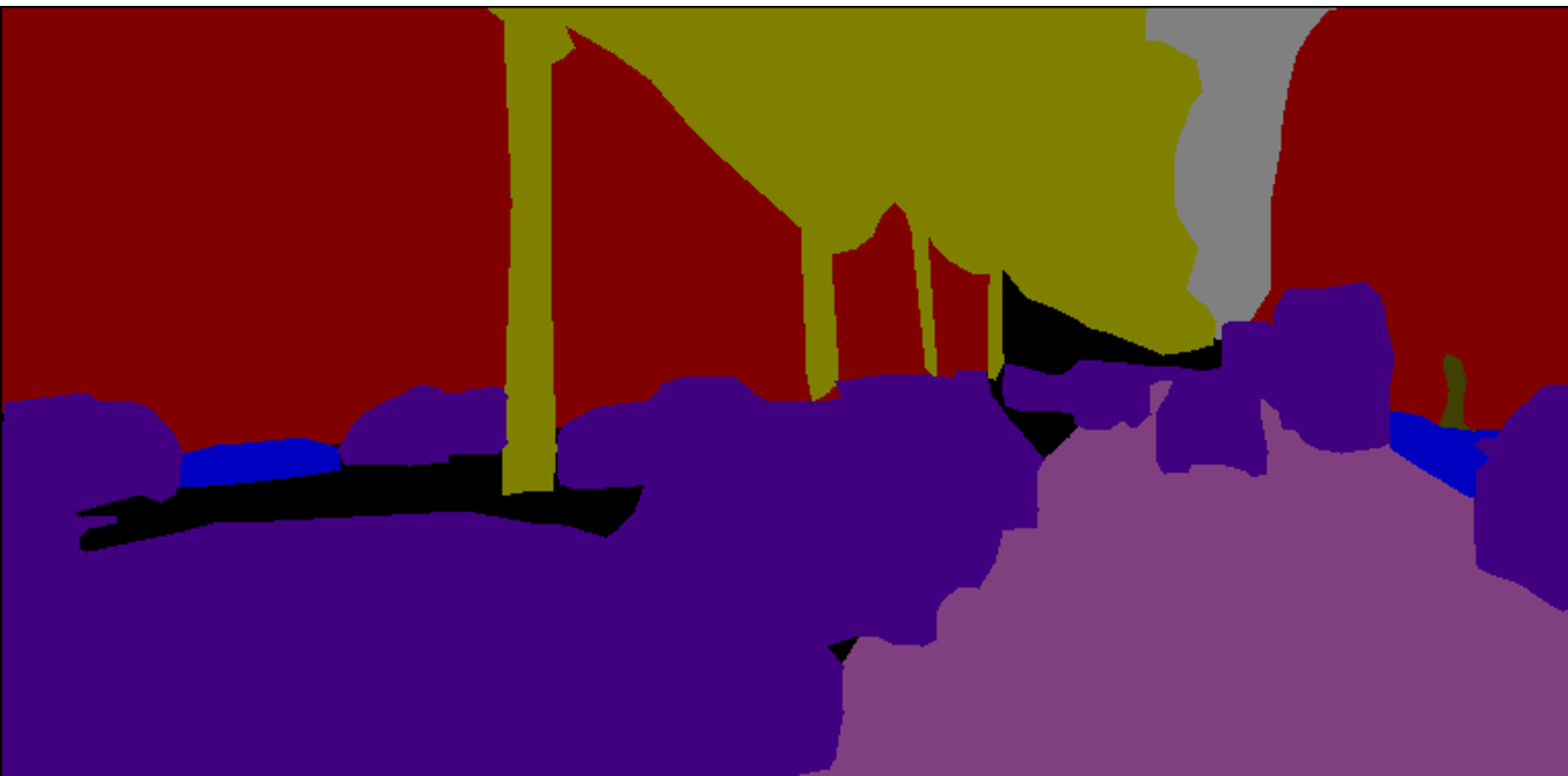} &
				\includegraphics[width=0.10\textwidth]{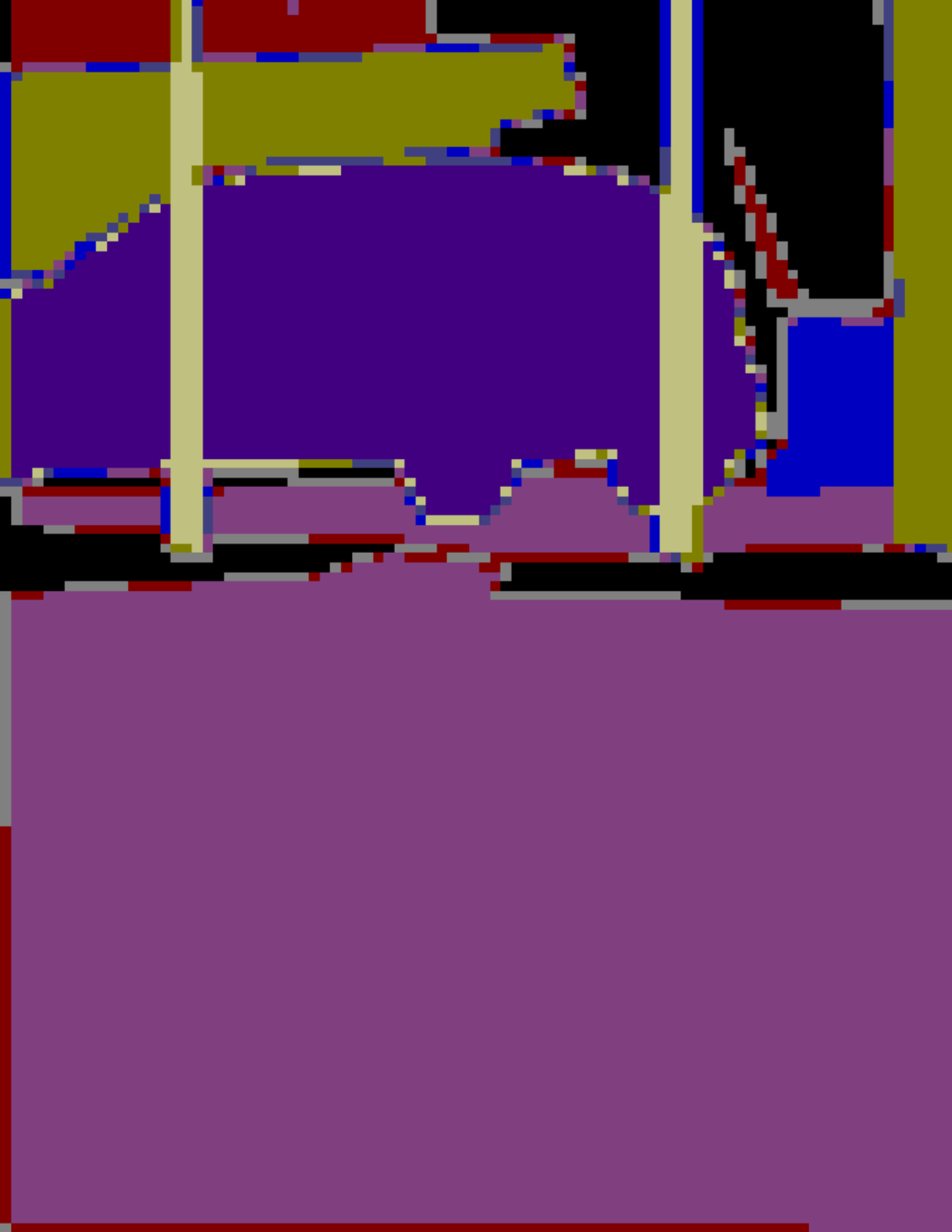}
				\includegraphics[width=0.10\textwidth]{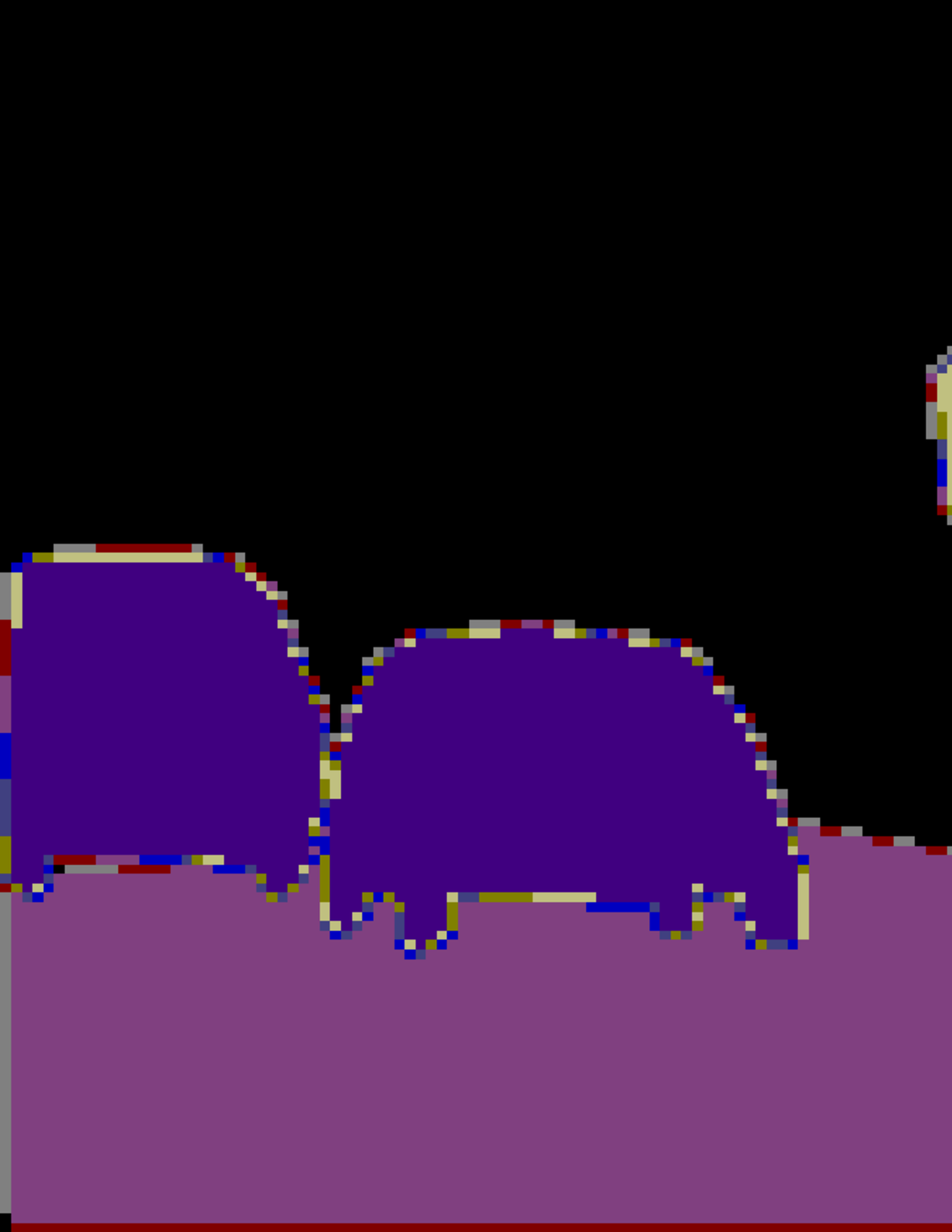} &
				\includegraphics[width=0.09\textwidth]{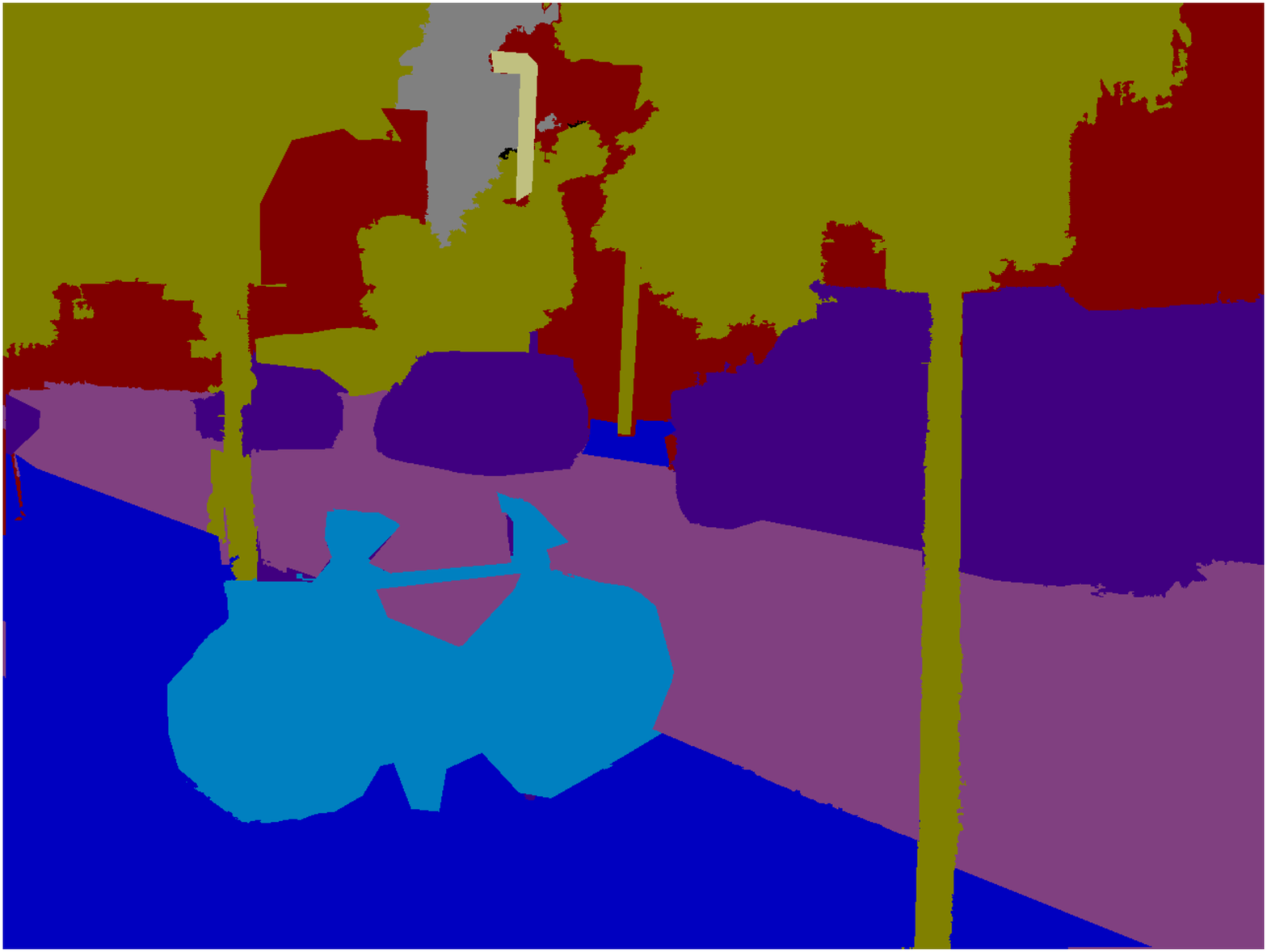}
				\includegraphics[width=0.09\textwidth]{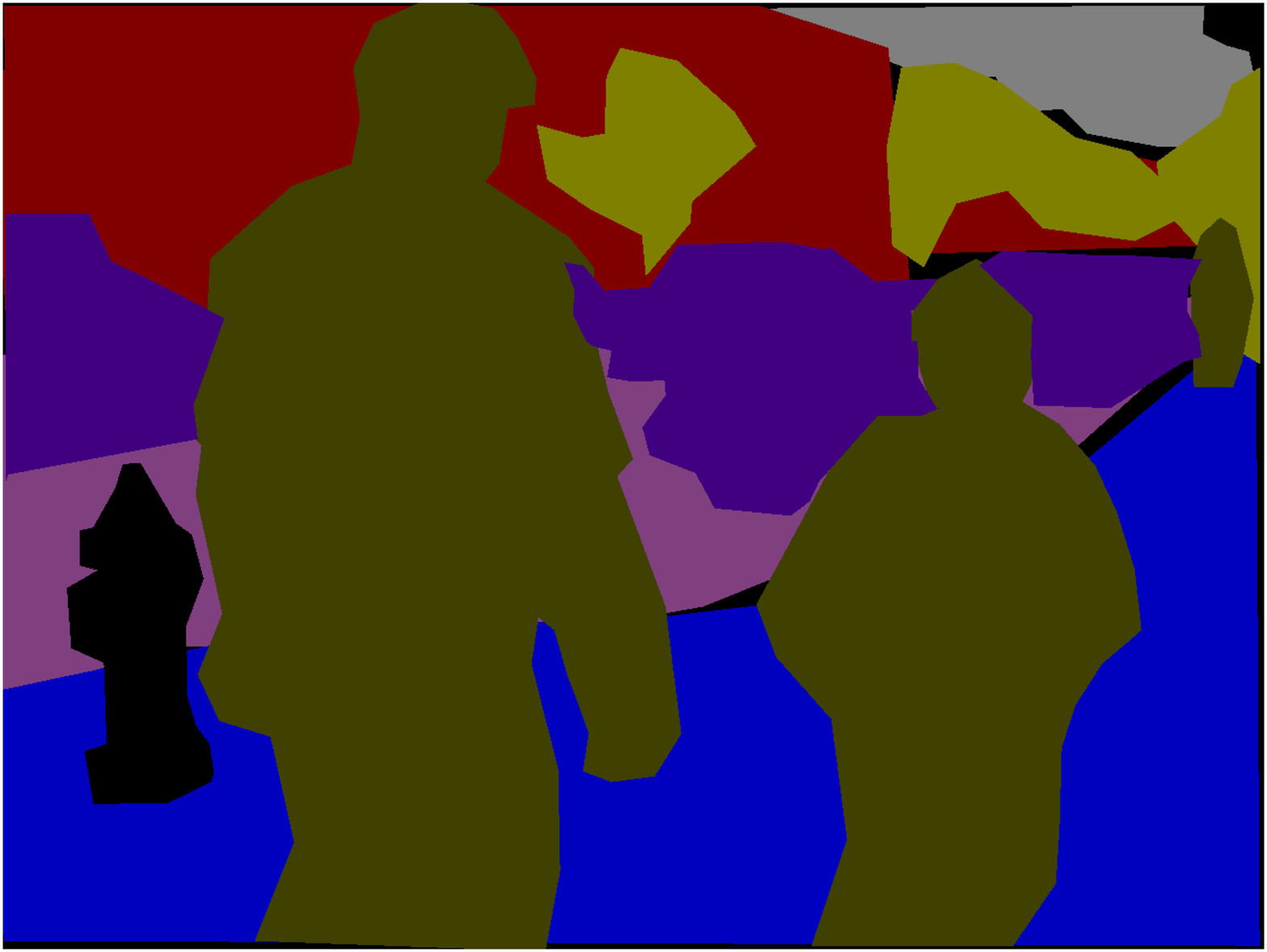} \\
			\textit{\# Images} & \textbf{600} (Cambridge, UK) & \textbf{547} (Karlsruhe, GER) & \textbf{942} (Various) & \textbf{3,547} (Boston, USA)\\
			%\textit{Location} & Cambridge, UK & Karlsruhe, GER & Various & Boston, MA, USA \\
			\textit{Used for} & Training & Training & \textbf{\textcolor{ForestGreen}{Test}} & \textbf{\textcolor{ForestGreen}{Test}} \\ %\bottomrule
			%& & & & \\
			\toprule
			\textbf{SPARSE:} & *ETH-RMPTMP (2009)~\cite{Ess09pami} & *GTSRB (2013)~\cite{Houben13ijcnn} & M-COCO (2014)~\cite{Lin14eccv} & *KITTI-O (2012)~\cite{Geiger13ijrr} \\ \midrule
			\textit{Example(s)} &
				\includegraphics[width=0.09\textwidth]{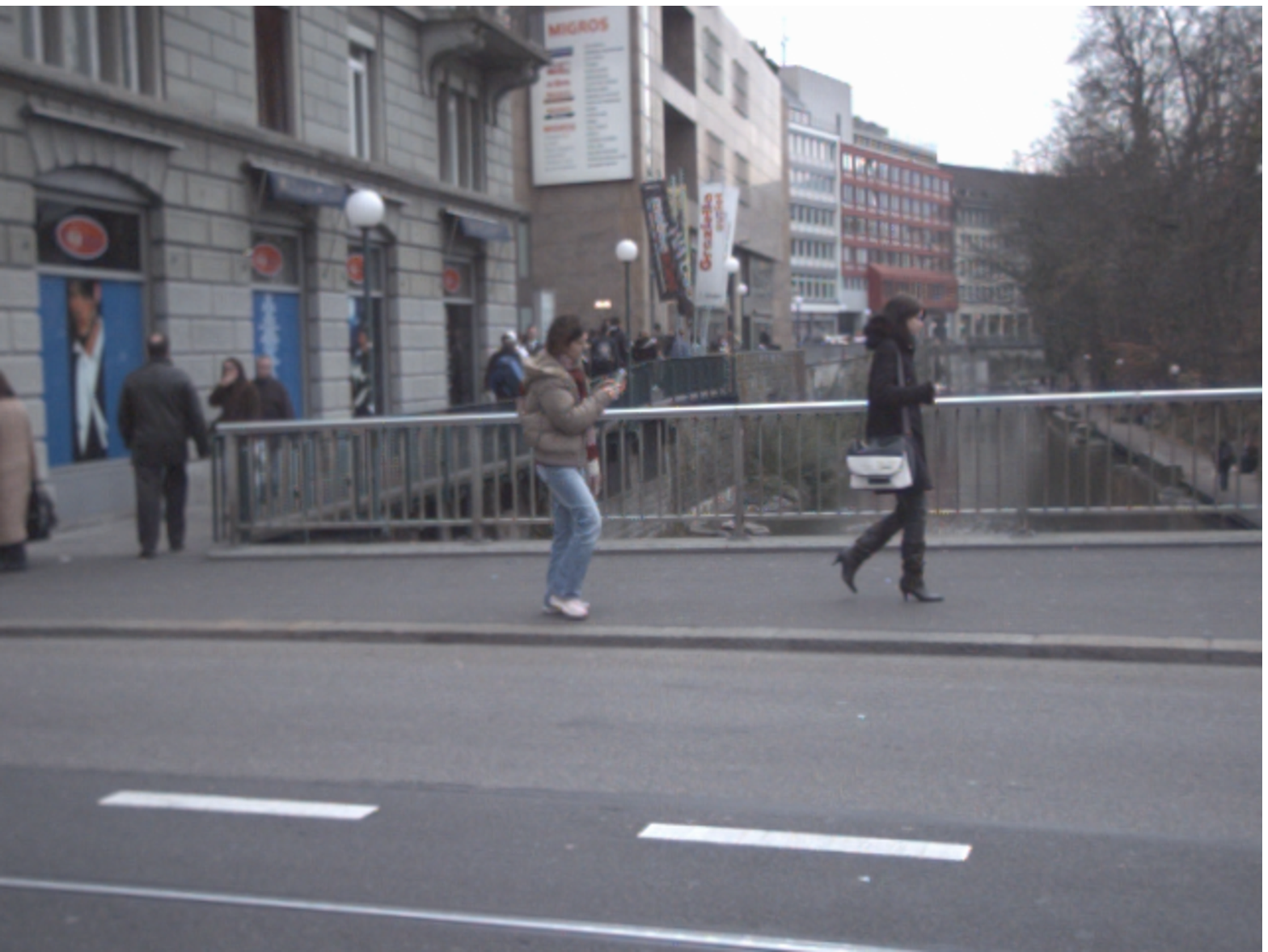}
				\includegraphics[width=0.09\textwidth]{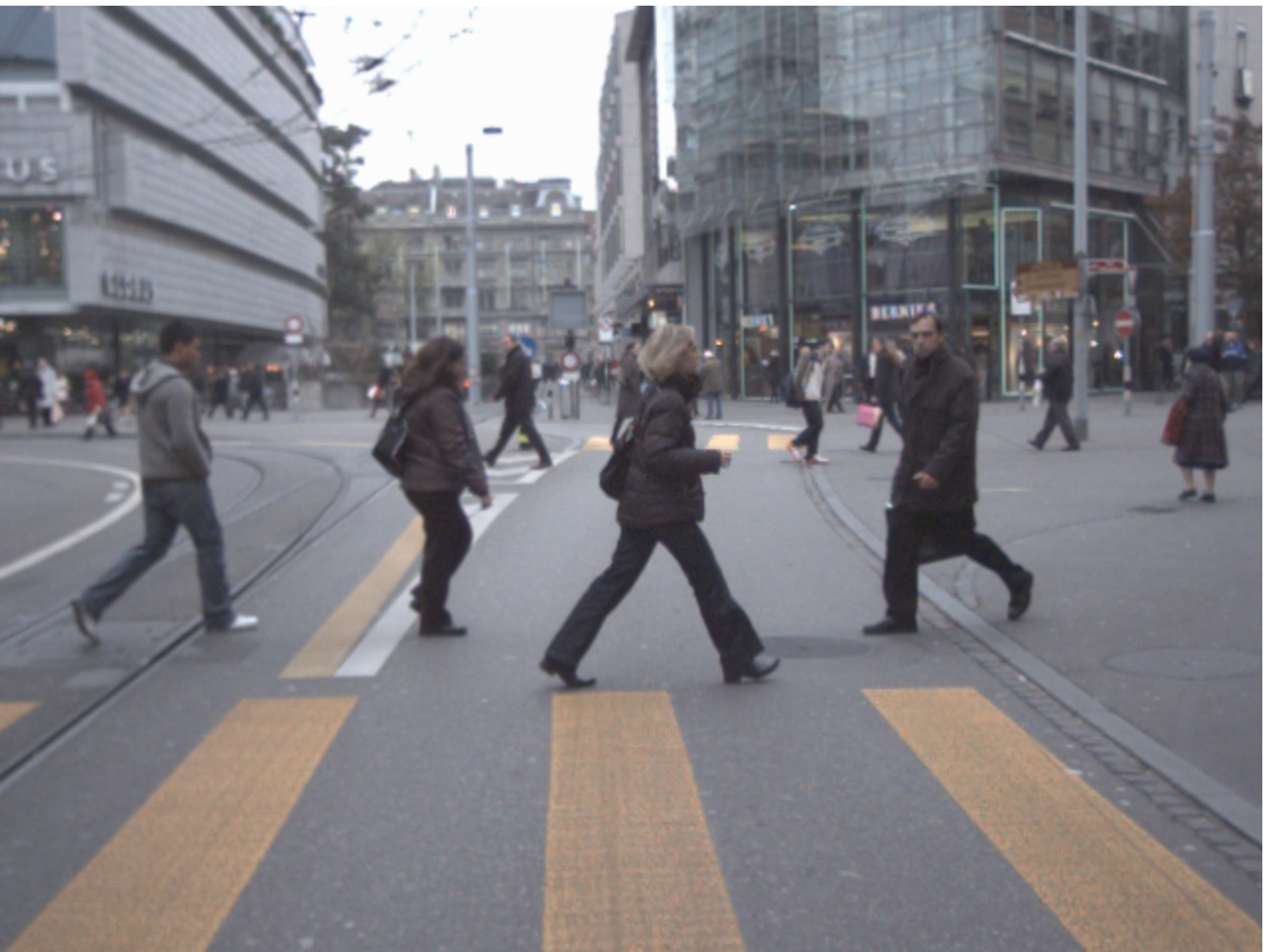} & 
				\includegraphics[width=0.09\textwidth]{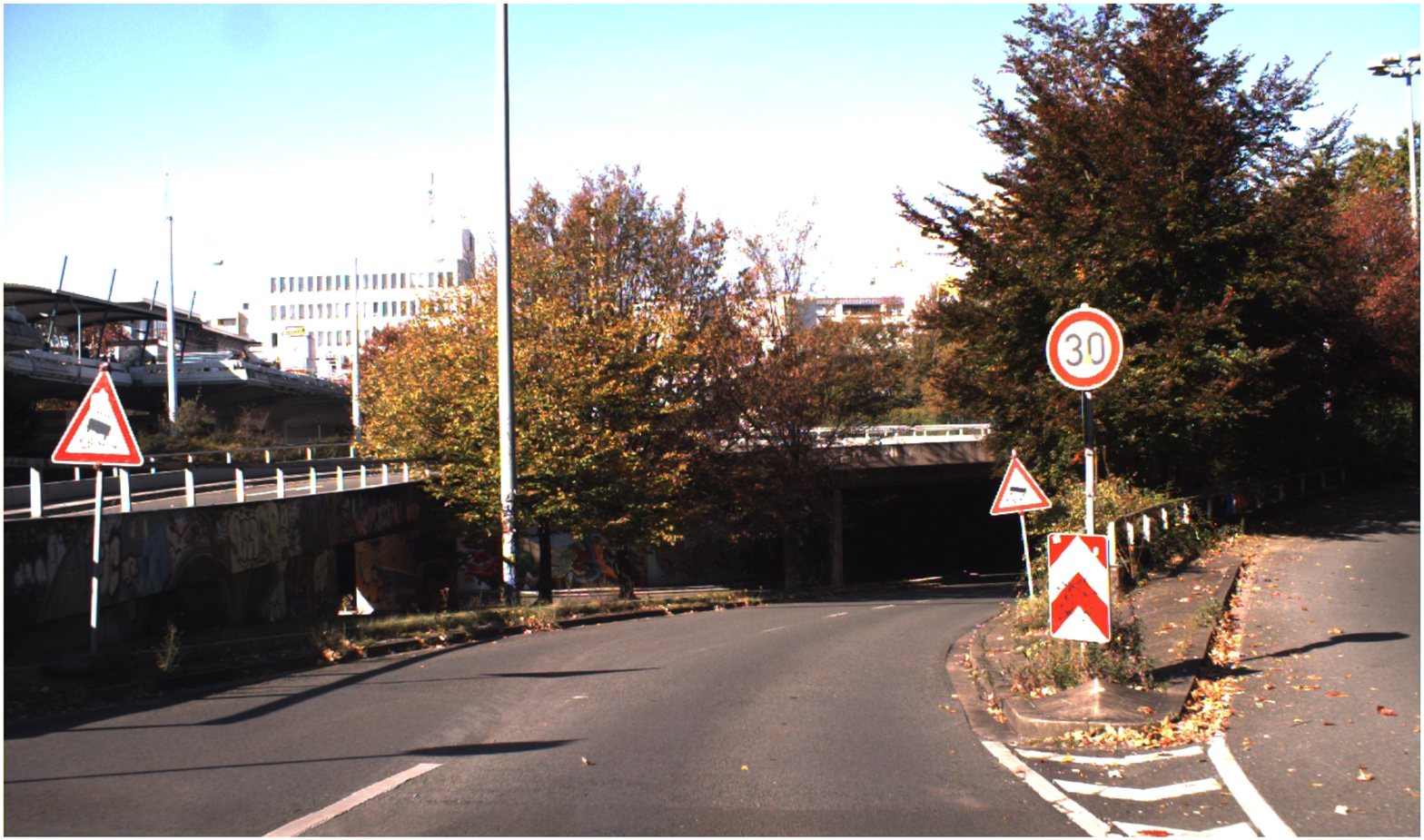}
				\includegraphics[width=0.09\textwidth]{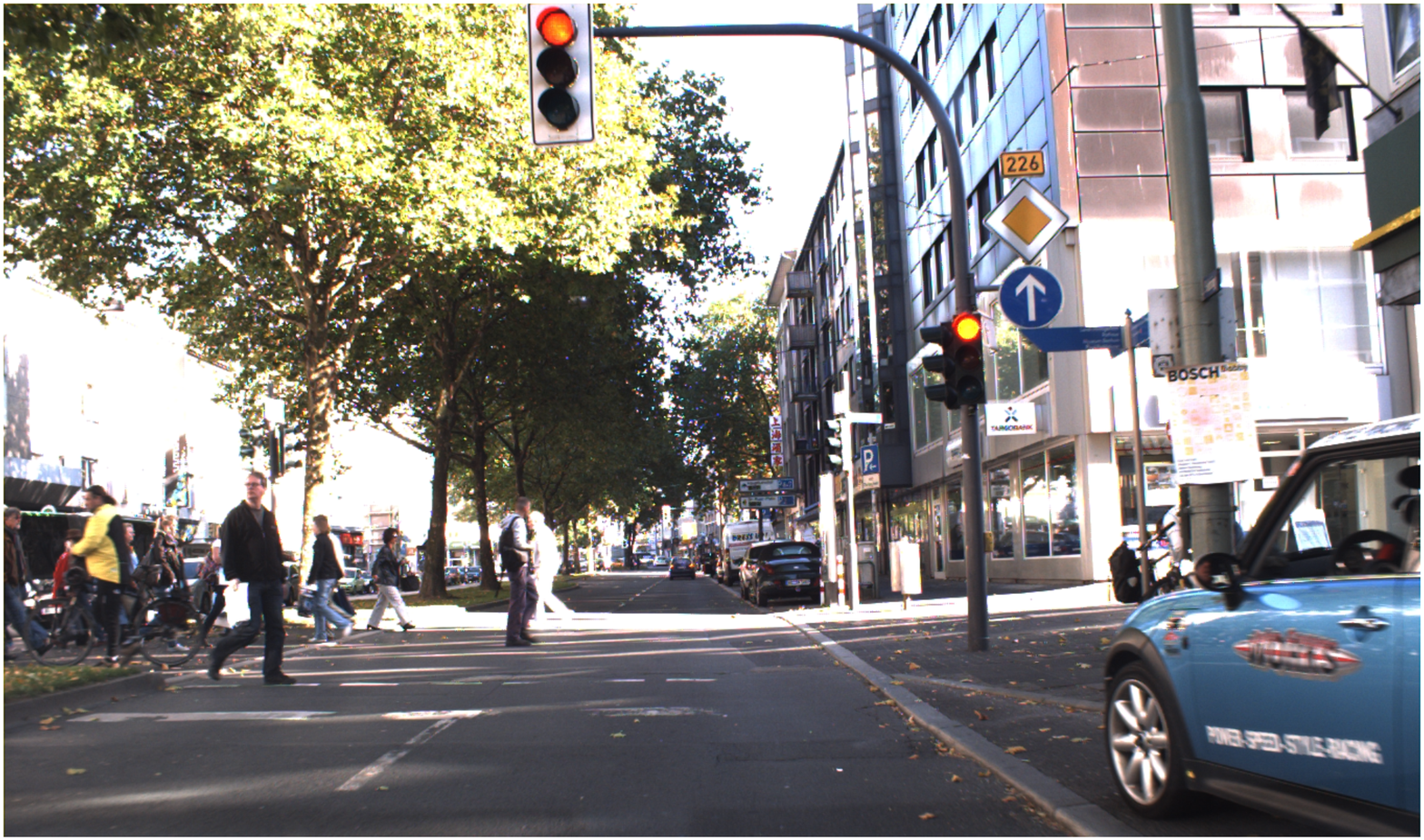} &
				\includegraphics[width=0.10\textwidth]{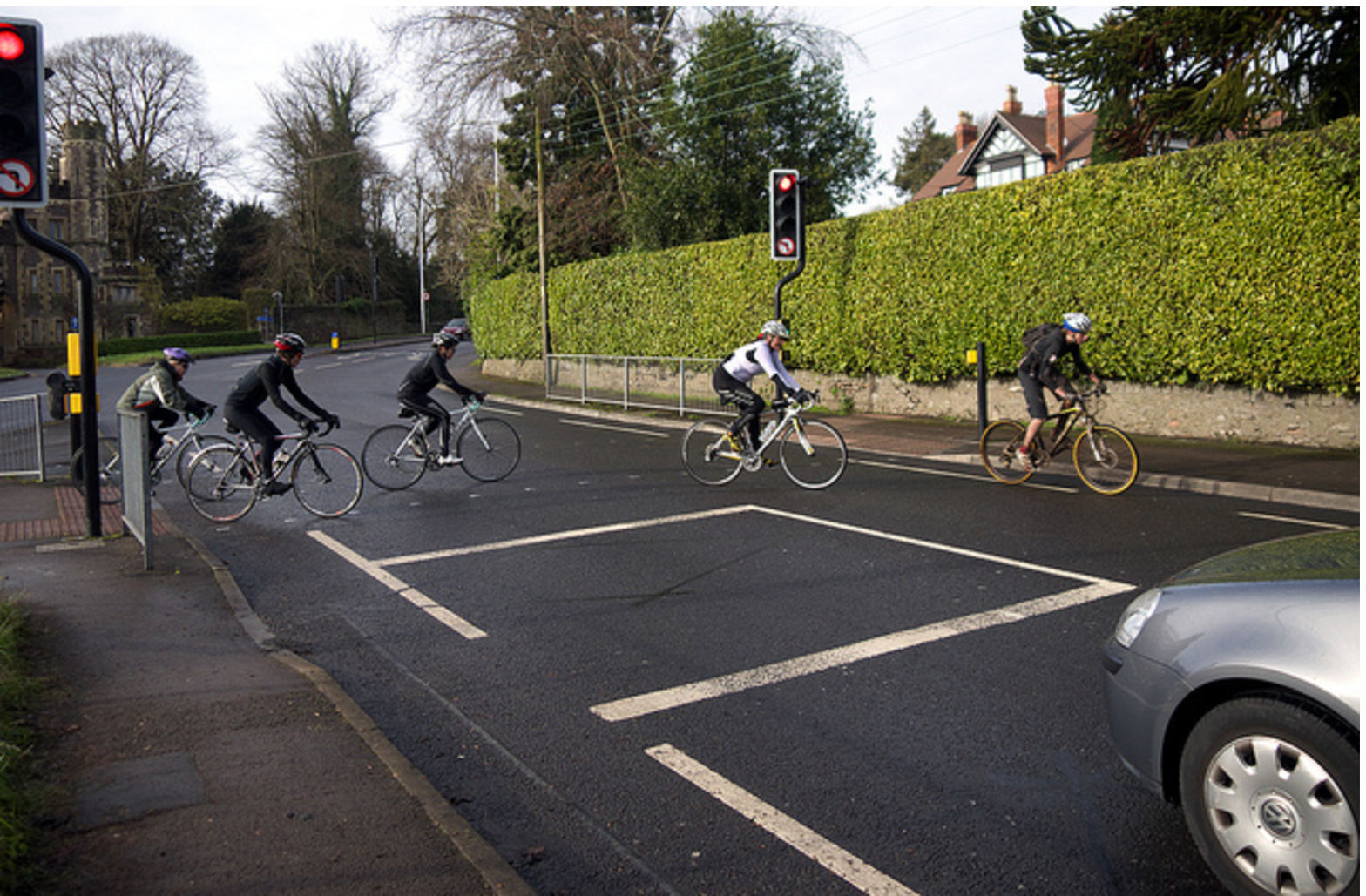}
				\includegraphics[width=0.10\textwidth]{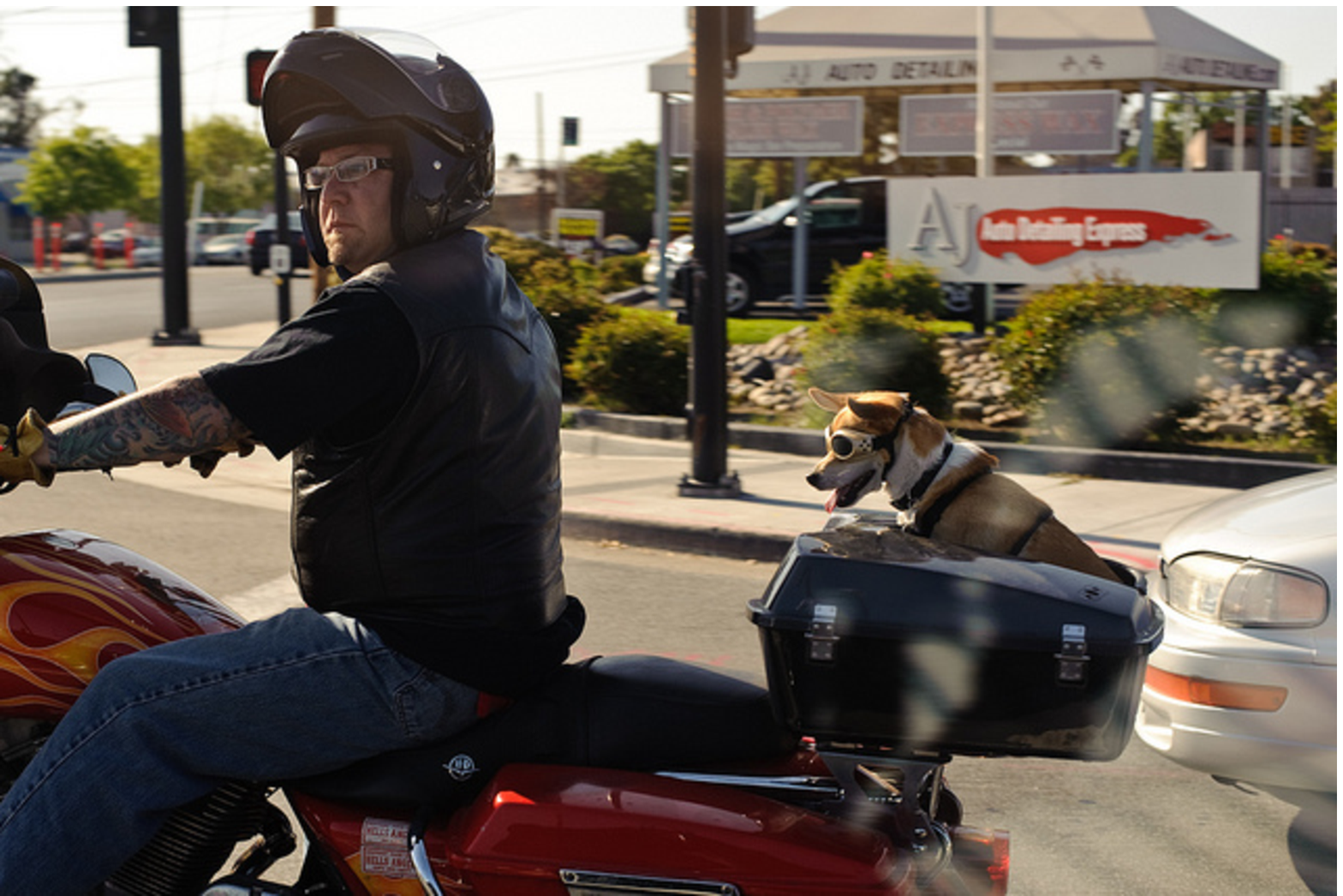} &
				\includegraphics[width=0.18\textwidth]{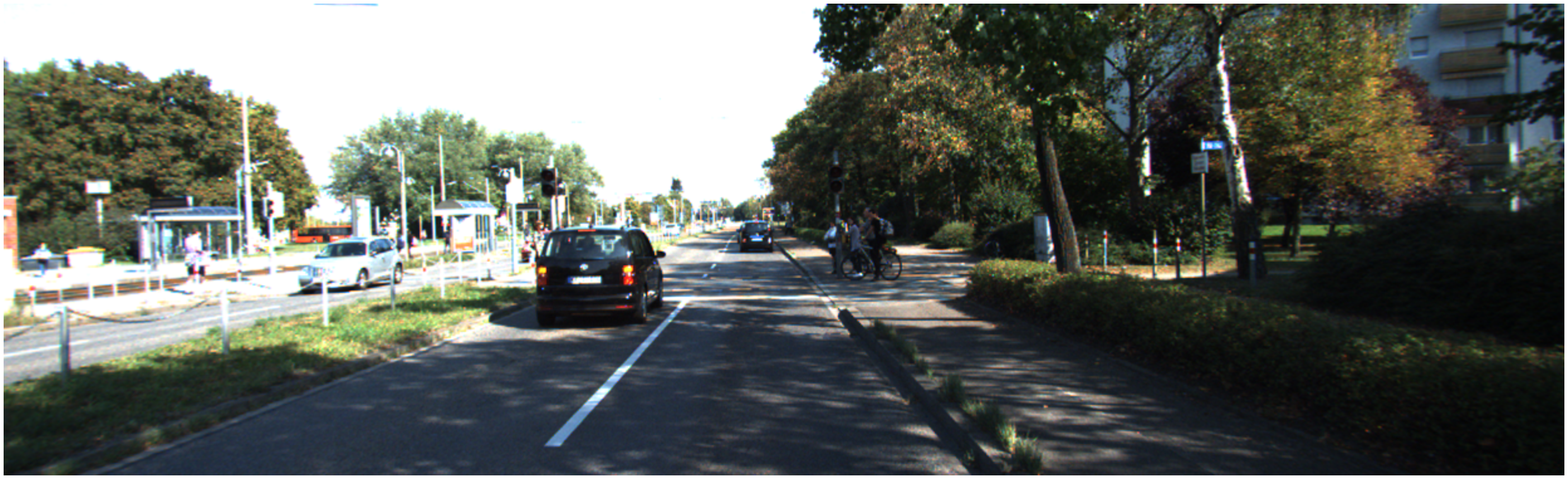} \\
			%\textit{Ground truth} &
			&
				\includegraphics[width=0.09\textwidth]{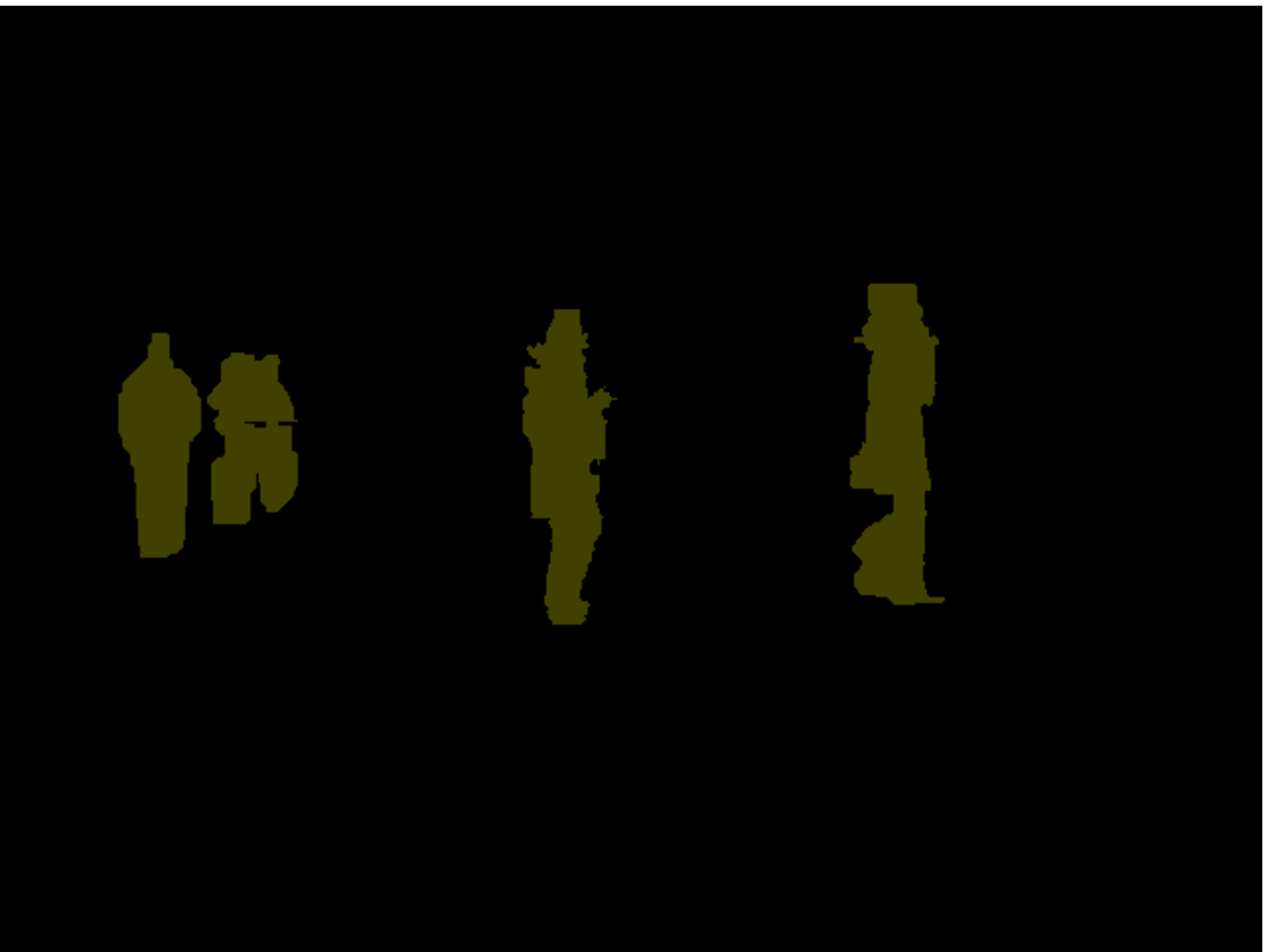}
				\includegraphics[width=0.09\textwidth]{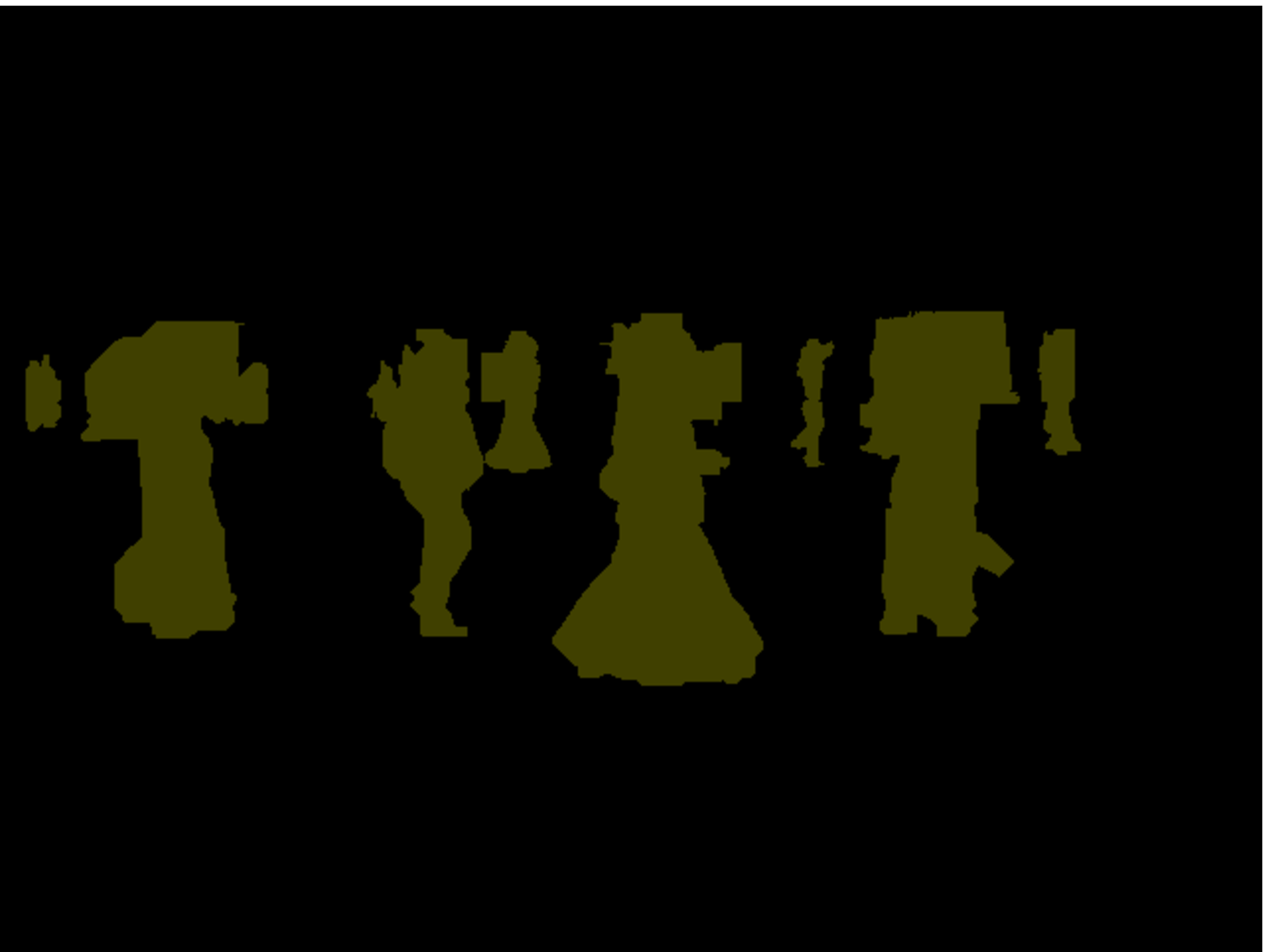} & 
				\includegraphics[width=0.09\textwidth]{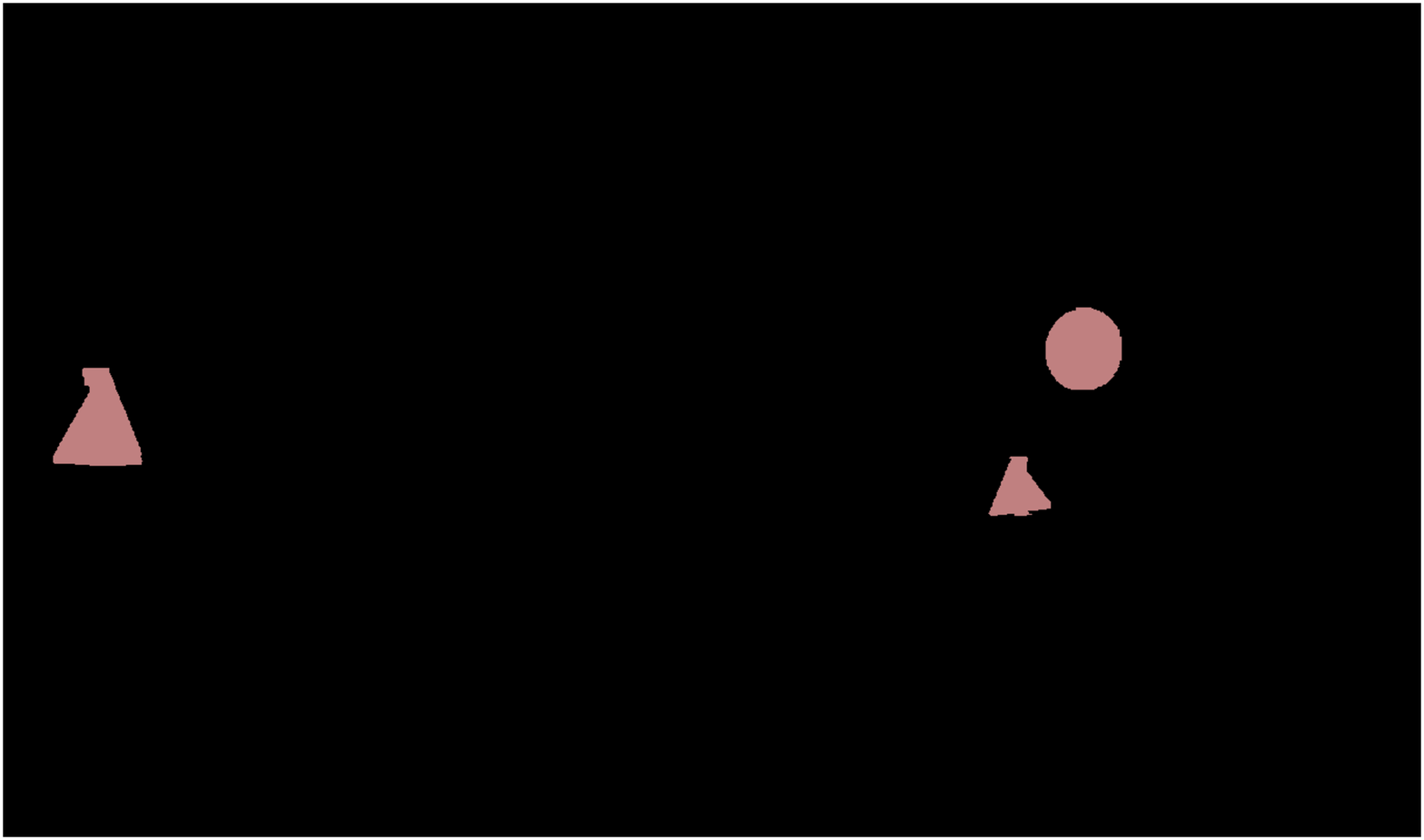}
				\includegraphics[width=0.09\textwidth]{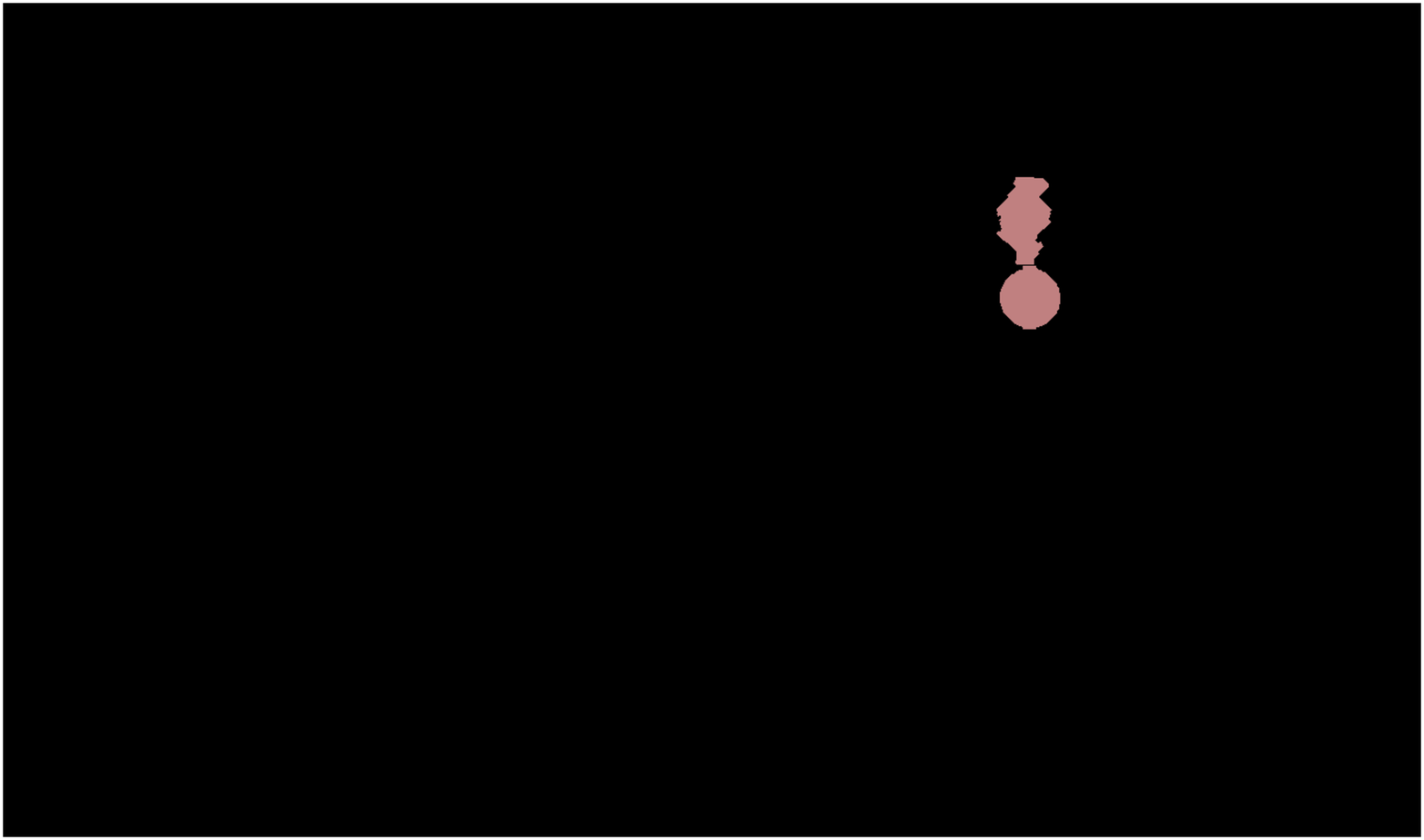} &
				\includegraphics[width=0.10\textwidth]{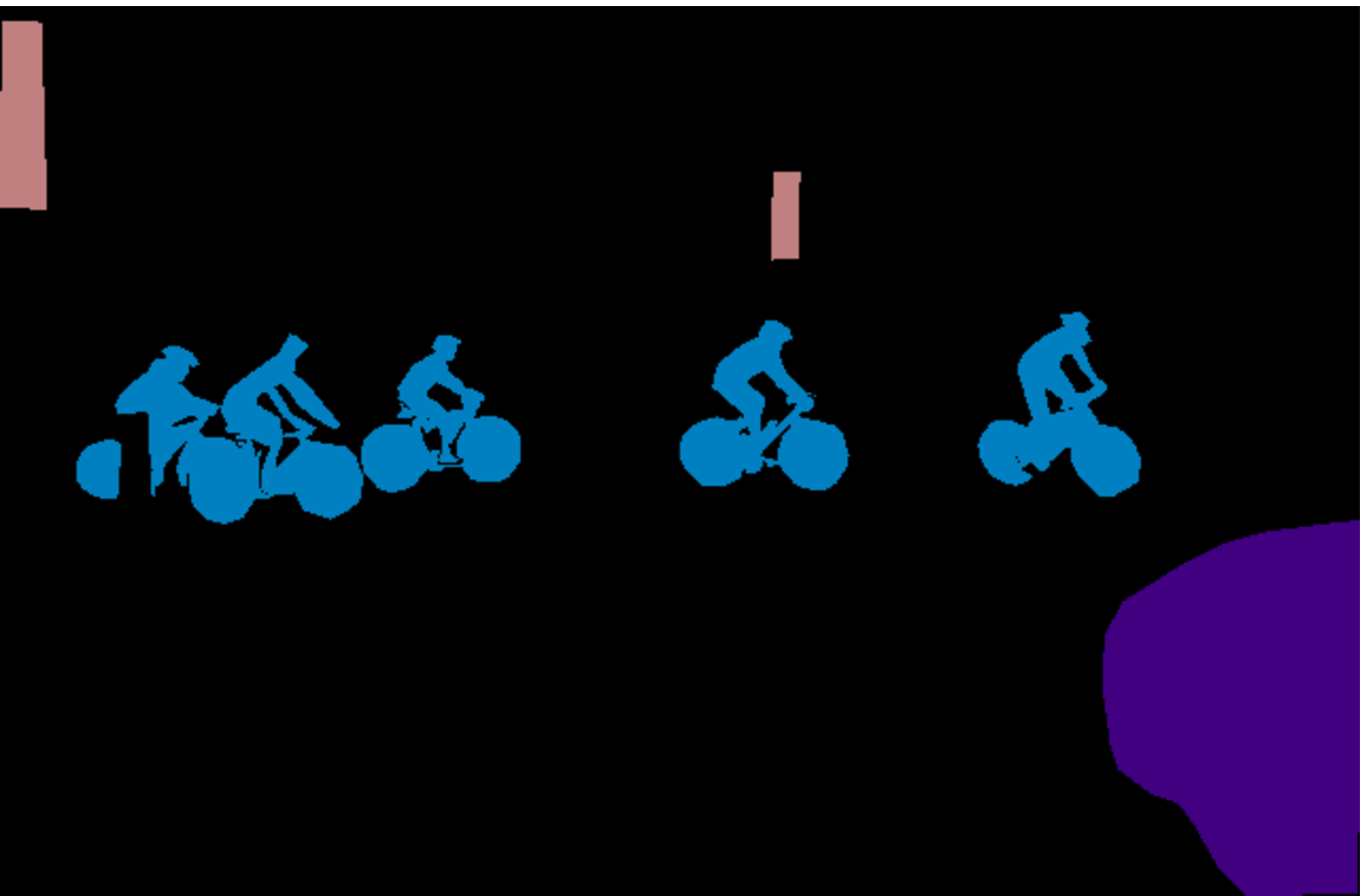}
				\includegraphics[width=0.10\textwidth]{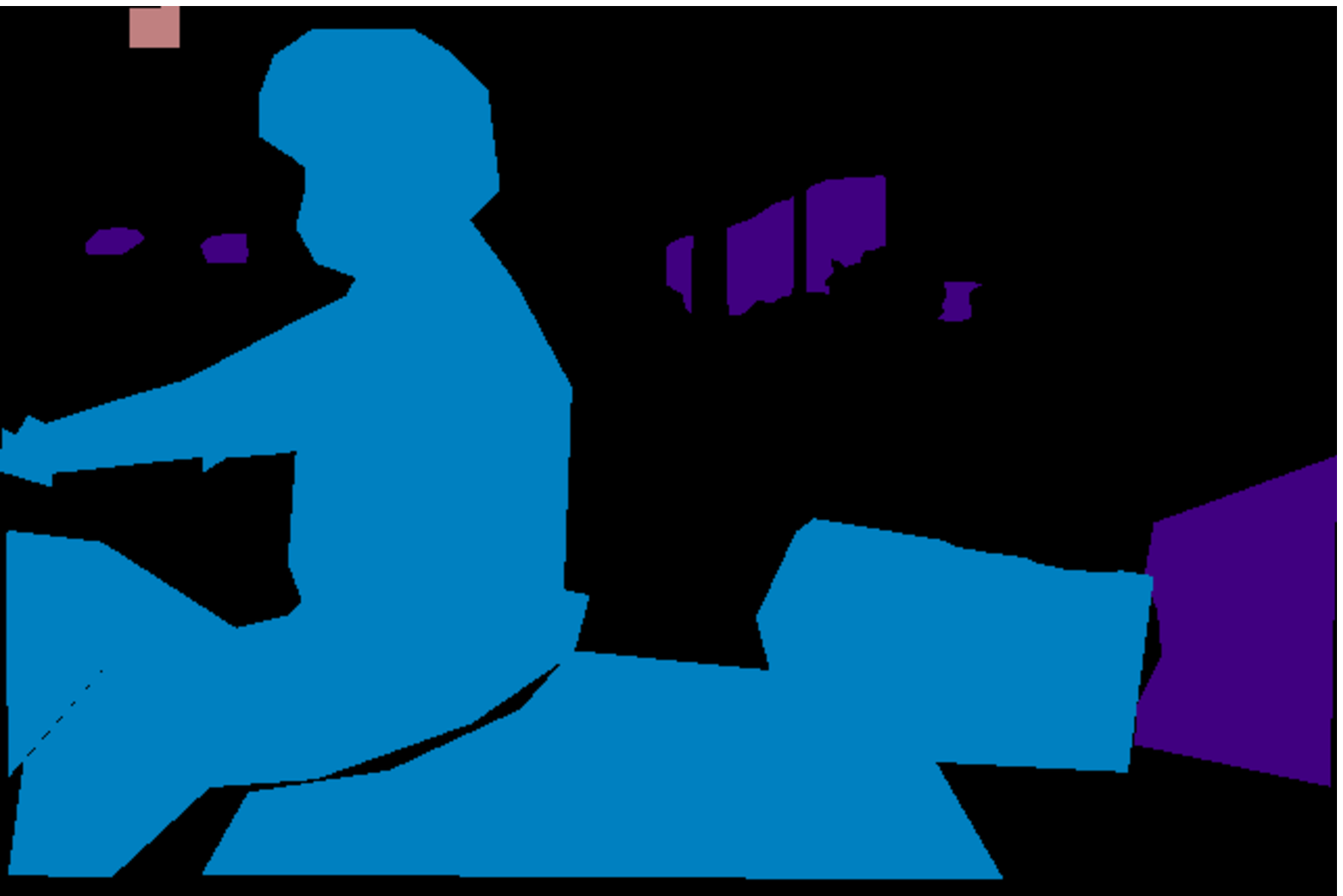} & \includegraphics[width=0.18\textwidth]{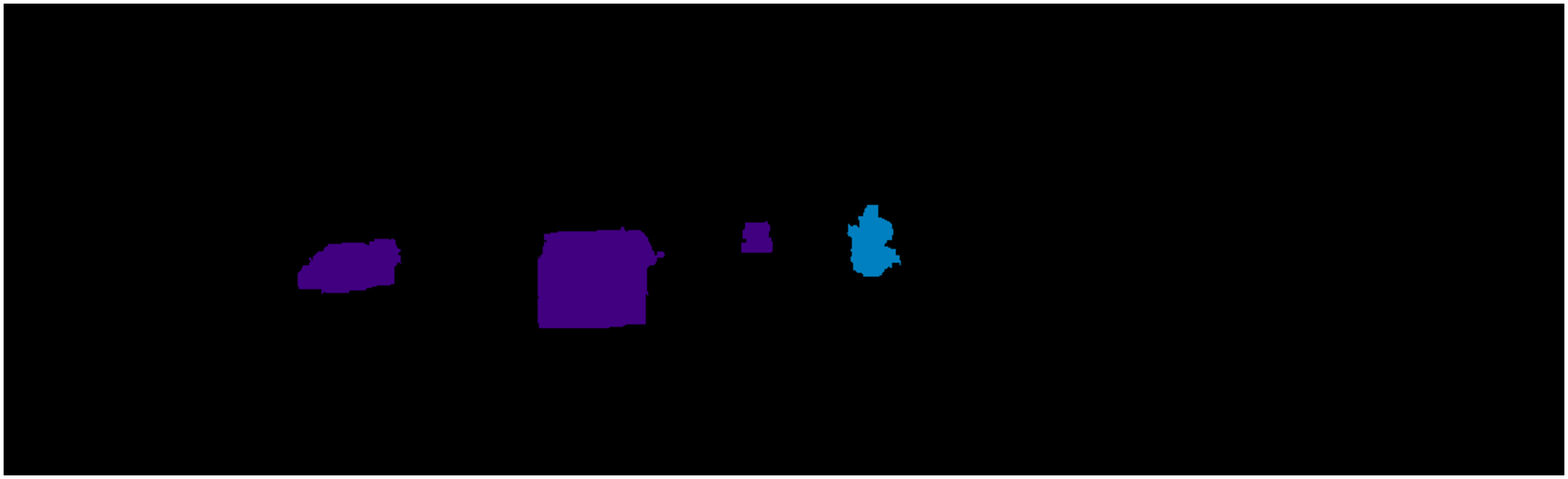} \\
			\textit{\# Images} & \textbf{14,056} (Zurich, CH)& \textbf{740} (Various, GER) & \textbf{3,262} (Various) & \textbf{7,481} (Karlsruhe, GER)\\
			%\textit{Location} & Zurich, SWIT & Various, GER & Various & Karlsruhe, GER \\
			\textit{Used for} & Training & Training & Training & Training \\ \bottomrule
%			& 2007 & Dense & 3,547 & Boston, MA, USA & 9 object categories\\ \midrule
%			 & 2008 & Dense & 600 & Cambridge, UK & Dash-cam, day and dusk \\ \midrule
%			& 2008 & Dense & 942 & Several places& \\ \midrule
%			 & 2013 & Dense & 547 & Karlsruhe, GER & \\ \midrule
%			 & 2009 & Sparse & 14,056 & Zurich, SUI & Pedestrian instances with b-box labels\\ \midrule
%			 & 2013 & Sparse & 7,481 & Karlsruhe, GER & Car/pedestrian/cyclist instances with b-box labels\\ \midrule
%			& 2013 & Sparse & 740 & BEL & Traffic signs with b-box labels\\ \midrule
%			 & 2014 & Sparse & 3,262 & Several places & Car/pedestrian/cyclist/road sign with polygon labels\\ \bottomrule
		\end{tabular}%
		}
		\vspace{-4mm}
	\caption{Constituent datasets of the Multi-Domain Road Scene Semantic Segmentation (MDRS3) dataset, containing dense and sparse labels. Datasets marked with * were upgraded from coarse labels or bounding boxes to pixel-wise annotations. The test set consists of two dense datasets (CBCL, Urban LabelMe) to better evaluate generalisation performance. Best viewed on screen.}%\TODO{check numbers.}}
	\label{fig:datasets}
	%\vspace{-4mm}
\end{figure*}

Acquiring data suitable for training road scene semantic segmentation is expensive and time-consuming. The process of densely labelling an image with 10-20 classes can take up to 30 minutes for a typical, cluttered perspective street-view image and so existing datasets tend to be relatively small. In addition, datasets are often confined to localised geographic regions and trained and tested on in isolation. In our work we consider using numerous datasets to create one aggregate dataset, which we refer to as the Multi-Domain Road Scene Semantic Segmentation dataset (MDRS3), to take advantage of all of the relevant training data available. 

%\subsection{Dataset composition}
%\label{sec:data:composition}
\paragraph{\bf Dataset composition.}
Fig.~\ref{fig:datasets} details the constituent datasets of our MDRS3 dataset.
We consider popular road scene semantic segmentation datasets with dense pixel-wise annotations such as 
CamVid~\cite{Brostow09prl,Brostow08eccv} and KITTI Semantic~(KITTI-S)~\cite{Geiger13ijrr,Kundu14eccv,Ros15wacv}.
However, as shown in Table~\ref{tab:class_distribution}, these dense datasets contain a large imbalance in the frequency of occurrence of various classes: structural classes such as road, sky or building are several orders of magnitude more frequent than important non-structural classes such as cars, pedestrians, road-signs or cyclists.
To boost the recognition of the latter, we include specific detection and recognition datasets where annotations are available in the form of bounding-boxes or segmentation masks: KITTI Objects (KITTI-O)~\cite{Geiger13ijrr}, a filtered set of Microsoft COCO (M-COCO)~\cite{Lin14eccv} containing pedestrians, cyclists, road signs and cars in urban environments, ETH Robust Multi-Person Tracking from Mobile Platforms (ETH-RMPTMP)~\cite{Ess09pami} for pedestrians and the German Traffic Sign Recognition Benchmark (GTSRB)~\cite{Houben13ijcnn} for road signs. 
The distribution of classes for our MDRS3 train and test sets (final two rows of Table~\ref{tab:class_distribution}) illustrate how training data in our dataset includes many more instances of important rare classes compared to existing dense datasets. 

\vspace{-2mm}
\begin{table*}[ht]
	\centering
	\caption{Class distribution (\% of total pixels) for the MDRS3 dataset constituents and test/train splits. The ``void" class has been removed for clarity.}\vspace{-3mm}
	\label{tab:data1}
	\resizebox{\textwidth}{!}{%
		\scriptsize
		\tabcolsep=0.21cm
		\begin{tabular}{@{}lr|rrrrrrrrrrrr@{}}
			\toprule
			          & & 
			          \includegraphics{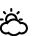} & \includegraphics{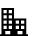} & \includegraphics{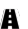} & \includegraphics{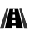} & \includegraphics{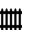} & \includegraphics{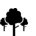} & \includegraphics{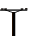} & \includegraphics{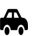} & \includegraphics{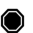} & \includegraphics{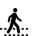} & \includegraphics{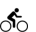}\\
			Dataset    & \# Image & sky & build. & road & side. & fence & veg. & pole & car & sign & ped. & cycl.\\ \midrule

			CamVid~\cite{Brostow09prl,Brostow08eccv} & 600 & 15.7 & 24.4 & 33.4 & 6.2 & 2.6 & 11.4 & 0.4 & 4.8 & 0.5 & 0.4 & 0.5 \\ \midrule
			KITTI-S~\cite{Geiger13ijrr} & 547 & 6.2 & 25.9 & 17.2 & 7.0 & 3.7 & 28.7 & 0.5 & 9.9 & 0.4 & 0.2 & 0.2 \\ \midrule
			*U-LabelMe~\cite{Russell08ijcv} & 942 & 13.2 & 39.9 & 19.1 & 8.1 & 0.3 & 11.1 & 0.5 & 5.8 & 0.3 & 1.1 & 0.5 \\ \midrule
			*CBCL~\cite{Bileschi07cbcl} & 3,547 & 5.4 & 26.4 & 28.2 & 6.9 & 0.7 & 17.9 & 1.3 & 11.8 & 0.3 & 0.8 & 0.2 \\ \midrule
			*ETH-RMPTMP~\cite{Ess09pami} & 14,056 & - & - & - & - & - & - & - & - & - & 100.0 & - \\ \midrule
			*GTSRB~\cite{Houben13ijcnn} & 740 & - & - & - & - & - & - & - & - & 100.0 & - & - \\ \midrule
			M-COCO~\cite{Lin14eccv} & 3,262 & - & - & - & - & - & - & 1.0 & 63.7 & 11.4 & 16.6 & 7.3 \\ \midrule
			*KITTI-O~\cite{Geiger13ijrr} & 7,481 & - & - & - & - & - & - & - & 90.7 & - & 7.4 & 1.9 \\ \midrule	\midrule
			MDRS3-Train & 26,686 & 5.4	& 12.1 & 12.5 & 3.1 & 1.5 & 9.2 & 0.5 & 36.6 & 3.4 & 13.2 & 2.5 \\ \midrule
			MDRS3-Test & 4,489 & 10.0 & 34.4 & 22.8 & 7.6 & 0.5 & 14.0 & 0.8 & 8.3 & 0.3 & 1.0 & 0.3 \\
			 \bottomrule
				\end{tabular}%
	}
	\label{tab:class_distribution}
\end{table*}

%#
\setlength{\textfloatsep}{10pt plus 1.0pt minus 12.0pt}
\setlength{\floatsep}{10pt plus 1.0pt minus 12.0pt}
\setlength{\intextsep}{12.0pt plus 2.0pt minus 8.0pt}
\vspace{-9mm}

%\subsection{Refinement of bounding box annotations}
%\label{sec:data:refinement}
\paragraph{\bf Refinement of sparse annotations.}
For constituent datasets where annotations are provided in the form of bounding-boxes (marked with an asterisk in Table~\ref{tab:class_distribution}), we perform refinement to pixel-wise annotations by adopting a similar GrabCut-based approach to~\cite{Papandreu15iccv}.
For the CBCL dataset, which is labelled with polygonal bounding-boxes for 9 object categories and contains many void areas, we enlarge the category set to 11 and extend existing labels to missing areas using a CRF classifier~\cite{Ladicky10eccv}. We provide further detail of this process in the accompanying supplementary material.

 %for obtaining semantic labels from bounding box annotations.
%we use the GrabCut algorithm~\cite{Rother04sg} to segment the foreground object pixels from the background. We constrain the central area of the bounding box ($\alpha \%$ of pixels within the box) to be foreground, while the pixels 
%outside the bounding box of the object of interest and the pixels inside the remaining bounding boxes of the other annotated objects in the image are constrained as background.
% We perform additional refinement on the CBCL dataset, which consists of 3,547 images with polygonal bounding boxes for 9 object categories (car, pedestrian, bicycle, building, tree, sky, road, sidewalk and store). It was  created for object detection and contains missing labelled objects, ambiguous labelling situations and objects that are not tightly bound within the polygon annotations. We re-labelled and extended the categories for 584 of the images containing dense pixel-wise annotations and produced semi-supervised annotations for the remaining 2,963 images by applying a CRF classifier~\cite{Ladicky10eccv} combined with the polygon-based ground truth.

%\subsection{Test dataset}
%\label{sec:data:test}
\paragraph{\bf Test dataset.}
For evaluation, we maintain a separation between datasets used for training and testing. We use a combination of different domains with 
dense and sparse annotations for training, while for testing we use two separate datasets with dense pixel-wise annotations: a new subset of the LabelMe dataset~\cite{Russell08ijcv} with urban images from different cities, referred here to as Urban LabelMe (U-LabelMe) and a processed subset of the CBCL StreetScenes Challenge 
Framework~\cite{Bileschi07cbcl}. 
These two datasets are more challenging compared to CamVid and KITTI, containing a larger variety of scenarios with different viewpoint and illumination conditions (compared to the forward-looking camera viewpoint in CamVid and KITTI). 
%(ii) the distribution of the frequency of the classes between structural and non-structural categories is more balanced than in CamVid or KITTI. 
Our test dataset thus provides a better measure of the generalisation performance of the trained network at test time, especially compared to the common practice of using subsets of the same sequence for training and testing.
% CNN Architecture and Learning Scheme
\section{Network Architectures for Semantic Segmentation}
\label{sec:networks}
\vspace{-2mm}
We consider two DN architectures and the trade-off they achieve between task performance and memory footprint. The two selected state-of-the-art networks are: the fully convolutional network (FCN)~\cite{Long15cvpr} and the DeconvNet~\cite{Noh15iccv}. We do not consider models that are extended with a CRF such as~\cite{Zheng15iccv}, since such extensions do not alter the intrinsic model capacity and smoothing can be added as a post-processing step if desired.

\begin{figure*}[t!]
\centering
\includegraphics[width=\textwidth]{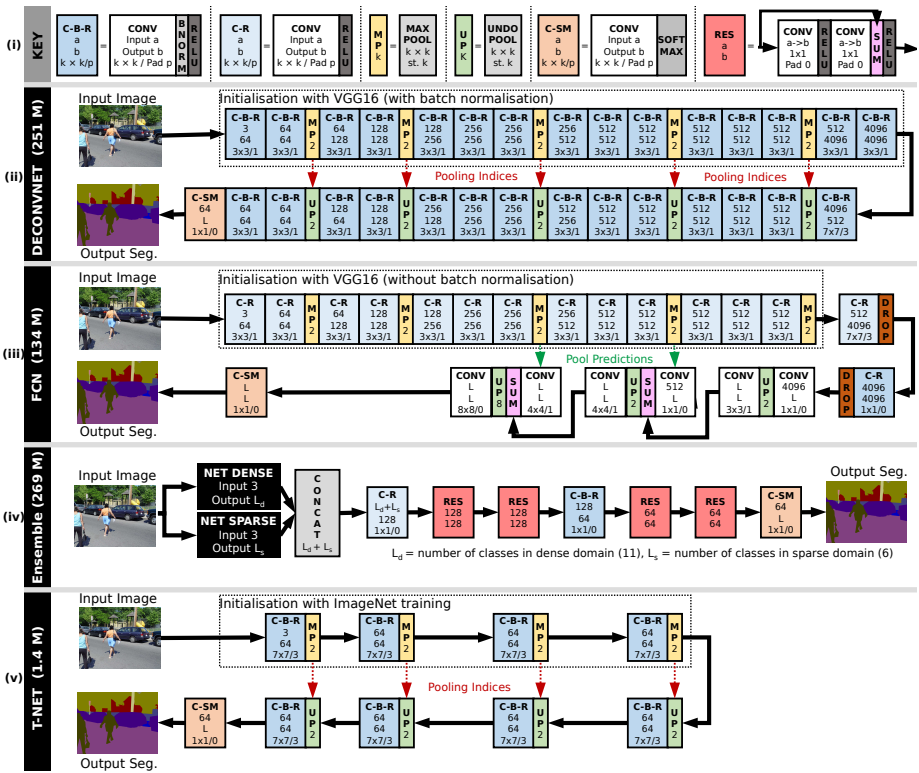}\\
\vspace{-4mm}
\caption{Summary of network architectures. From top to bottom: (i) a key to basic building blocks; (ii) DeconvNet~\cite{Noh15iccv} (251M parameters); (iii) FCN~\cite{Long15cvpr} (134M parameters); (iv) our constrained Target Net or T-Net (1.4M parameters); and (v) our dense/sparse FCN ensemble or chosen S-Net (269M parameters). All convolutions assume stride 1 unless otherwise specified. Descriptions of these networks can be found in sections~\ref{sec:networks} and \ref{sec:training}.}
\label{fig:cnn_archs}
\end{figure*}

The DeconvNet and FCN architectures are shown in Fig.~\ref{fig:cnn_archs}(ii)-(iii), respectively. Both DNs expand on
VGG-16~\cite{Simonyan15iclr}, but the DeconvNet is much deeper than FCN (75\% more parameters), making it harder to train and unsuitable for embedded 
applications. The depth of these networks is justified for the task of semantic segmentation of general scenes (which contain thousands of classes of objects), but shallower 
networks may suffice for constrained urban environments. Another difference between these architectures is in the 
upsampling philosophy: the FCN combines outputs of different layers to achieve better localization accuracy, while the DeconvNet stores 
pooling indices to re-use later for guiding feature map upsampling. This last strategy has been shown to improve localization accuracy for different problems~\cite{Zeiler10cvpr}.

%We opted to use FCN as the base architecture for our Source Network (S-Net), due to its good performance and the ease of training. For the Target-Net we chose a simplification of DeconvNet, similar in spirit to works such as~\cite{Badrinarayanan15arxiv}. 
\vspace{-4mm}
\subsection{Source Network (S-Net) Architecture}\label{sec:target}\vspace{-1mm}

The Source Network (S-Net) is selected by choosing the best possible performing network, disregarding memory or computational constraints. 
%In the context of this work, the S-Net must be understood as a reference network. We aim to replicate its modelling behaviour 
%using a simpler and smaller network -- the Target Net (T-Net). The S-Net is created without memory or computational constraints since it will only be used during training.
Our choice of S-Net consists of an ensemble of two FCN networks trained on different 
data modalities, \ie dense and sparse data modality---see Fig.~\ref{fig:cnn_archs}(iv)---, which was found to be the best performing unconstrained network as reported in section~\ref{sec:exps}. 
%This decision is motivated by the superior accuracy of the ensemble as reported in Section~\ref{sec:training}. 
Although this ensemble contains more parameters than a DeconvNet (269M versus 251M), it leads to better results and is faster to train, reason why we do not use DeconvNet and favour FCN-based approaches instead for S-Net. Further detail is contained in section~\ref{sec:training_ensemble}.

\vspace{-2mm}
\subsection{Target Network (T-Net) Architecture}\label{sec:production}\vspace{-1mm}
The T-Net, shown in Fig.~\ref{fig:cnn_archs}(v), consists of a model based on the pooling-unpooling principle of the 
DeconvNet~\cite{Noh15iccv} but simplified to suit an embedded system (similar to the ``basic" SegNet version 
proposed in~\cite{Badrinarayanan15arxiv}).

T-Net consists of 4 \textit{contraction} blocks, followed by 4 \textit{expansion} blocks, with a total of 
$1.4$ M parameters. This reduced size offers a good compromise between memory requirement and  performance. Contraction blocks serve to create a rich representation that allows for recognition as in standard 
classification CNNs. Expansion blocks are used to improve the localization and delineation of label 
assignments. Both contraction and expansion blocks use $7{\times}7$ kernels with a stride of 1 pixel and a fixed number of 64 feature maps. Batch normalization is added prior to ReLU to reduce internal covariate shift~\cite{Ioffe15icml} during training and improve convergence. Upsampling in expansion blocks is carried out by storing and retrieving pooling indices for current activations. This helps to produce sharp edges in the final output, avoiding blocky results~\cite{Noh15iccv}. A linear classifier performs the final label estimation at the pixel level. The choice of 4 blocks is motivated by empirical analysis, offering the best trade-off between model compactness and good performance (see supplementary material).

%T-Net consists of 8 blocks, \ie 4 \textit{contraction} blocks, followed by 4 \textit{expansion} blocks, all of them with a fixed number of 64 feature maps. The output of these blocks is connected to a softmax classifier as depicted in  Fig.~\ref{fig:cnn_pnet}. \textit{Contraction} blocks are on charge of learning general representations as in a classical CNN. The size of the input data is reduced in each block, which is comprised of a convolution (CONV), batch normalization (BNORM), ReLU and max pooling. \textit{Expansion} blocks recover the original size of the data and are on charge of improving localization of the different classes, \ie to perform accurate inference at the level of pixel respecting object boundaries. These blocks are comprised of unpooling, convolution,   batch normalization and ReLU layers. This new architecture, referred here as T-Net, having only $1.4$ M parameters, offers a good compromise between memory requirements, speed and  performance. 

\section{Training Strategies for DN architectures}\vspace{-2mm}
\label{sec:training}
In this section we describe the different approaches used to train our networks on the challenging MDRS3 dataset 
described in section~\ref{sec:datasets}. The approaches explored are (i) ``e2e"---standard end-to-end training over various subsets of the 
multi-domain training data; (ii) ``BGC"---which uses Balanced Gradient Contribution to generate stable gradient directions for end-to-end 
training; (iii) ``Flying-Cars"---dynamic domain adaptation of the sparse training data; and (iv) ``Ensemble"---the ensembling of models 
trained on separate domains. The performance of these methods is evaluated in section~\ref{sec:exps}.  

%In terms of training data, we assume the sets described in Section~\ref{sec:datasets}. For simplicity these sets have been split in two general modalities or domains: a first domain containing typical pixel-wise annotations that are dense, \ie, fine annotations covering all the image and containing background classes (we refer to this domain as dense); and a second domain containing noisy labels that are present just for some objects and represent a small percentage of the image (we refer to this domain as sparse). The main difference on the training methods is on how to exploit the information contained on these domains.

Each training strategy was initialised identically. Contraction blocks of the architecture under study were assigned the weights of classification networks pre-trained on ImageNet -- VGG-16~\cite{Simonyan15iclr} in the case of FCN and VGG-F~\cite{Chatfield14bmvc} in the case of T-Net. Adjustments to the shape of the weights were performed where dimensions do not match. Expansion blocks were initialized using the method of He \etal\cite{He15iccv}.

Optimisation was performed via standard backpropagation using Stochastic Conjugate Gradient Descent (S-CGD), endowed with a bounded line-search strategy and backtracking with Armijo's rule~\cite{LeNCLPN11}. To avoid overfitting, the number of line-search iterations was bounded to 3.%$T_\text{line-search} = 3$ in our case.
This proved to converge faster to good solutions than stochastic gradient descent without manual tweaking of learning rates.

%
%All the data is normalized using local contrast normalization~(LCN), followed by mean subtraction and magnitude re-scaling in the range $[-128, 
%128]$. Then, we initialize each of the network kernels following a random Gaussian distribution with $\mu=0$ and $\sigma = \sqrt{2/(H \times 
%	W \times \text{fan}_\text{out})}$~\cite{He15iccv}, where $H \times W$ and fan out are the spatial dimensions and the fan out of the kernel, 
%respectively. 

%All the trainings are carried out using Stochastic Conjugate Gradient (S-CG)~\cite{} endowed with a bounded line-search strategy based on the Armijo's method~\cite{}. This removes the need of adjusting learning rates and speeds up convergence, being therefore preferred over classical SGD
\vspace{-2mm}
\subsection{End-to-End Training (e2e)}\vspace{-1mm}
\label{sec:training_endtoend}

Our simplest training approach, end-to-end (e2e) training, consisted of standard mini-batch training on random samples (with replacement) from the mixed dense and sparse training set (\ie all data in). To achieve reasonable per-class accuracy, the use of weighted cross-entropy (WCE) as training loss was found to be essential. WCE re-scales the importance of each class, $l \in [1, \dots, L]$, according to its inverse frequency $f^{l}(\mathcal{X})^{-1}$ in the training data set $\mathcal{X}$, \ie:

\vspace{-2mm}
\begin{equation}%
\mbox{\normalsize\( %
\text{Loss}_\text{WCE}(x^n, y^n) = -\sum_{ijl}^{HWL} \omega(y^n_{ijl}) y^n_{ijl} \text{log}(\mathcal{F}(x^n, \theta))_{ijl},%
\)} %
\end{equation}
\noindent where $\mathcal{F}$ refers to the network, $x^n$, $y^n$ stand for the $n$-th training image and ground truth image respectively, and 
%$\omega$ is a function mapping from $[1, \dots, K] \rightarrow  W$, where $W \in \mathbb{R}^K$ is a vector of weights computed from the training images $\mathcal{X}$ as:
%\begin{equation}
%\mbox{\normalsize\( %
%W^k = \text{max} \left\lbrace \frac{f^k(\mathcal{X})^{-1}}{\sum_{i=1}^K f^{i}(\mathcal{X})^{-1}  %\text{ min} \{   f^{1}(\mathcal{X})^{-1}, \dots,  f^{K}(\mathcal{X})^{-1}\} }, \tau %\right\rbrace .%
%\)} %
%\end{equation}
the weighting function is given by
\begin{equation}
\mbox{\normalsize\( %
w(y_{ijl}^n) = \text{max} \left\lbrace \frac{f^l(\mathcal{X})^{-1}}{\sum_{i=1}^L f^{i}(\mathcal{X})^{-1}  \text{ min} \{   f^{1}(\mathcal{X})^{-1}, \dots,  f^{L}(\mathcal{X})^{-1}\} }, \epsilon \right\rbrace\)}, \text{ for } \epsilon = 10^{-5}.
\end{equation}
%\begin{equation}
%W^c = \text{max} \left\lbrace \frac{f^c(\mathcal{X})^{-1}}{\sum_{i=1}^C f^{i}(\mathcal{X})^{-1}  \text{ min} \{   f^{1}(\mathcal{X})^{-1}, \dots,  f^{C}(\mathcal{X})^{-1}\} }, \tau \right\rbrace .
%\end{equation}

In this way, WCE helped the networks to account for class frequency imbalances, a common phenomenon exposed in Table~\ref{tab:class_distribution}, which were otherwise observed to reduce a network's attention to rare but important classes such as pedestrians or bicycles during training.

End-to-end training was applied to learn separate models for the dense and sparse domains as well as a combined model on both data domains. However, when this approach is used naively on the combined data, we observed an unstable oscillatory behaviour of the objective and eventually divergence of the system. This phenomenon is due to the strong difference between the statistics of both distributions, which give rise to very noisy descent directions during optimisation. Thus, in order to exploit all the information available in both domains one needs to stabilize the training process, via alternatives such as those proposed in the following sections. 

%several issues depending on the architecture. For FCN there is an oscillatory effect that slows the convergence of the net down and makes it unstable. For T-Net this effect is magnified and makes the model produce oscillatory predictions with the most dominant sparse classes. In the last case, this behaviour has to do with batch normalization layers, that compute local statistics to normalize batches, and these statistics are severely affected by the discrepancies between distributions. In other words, this phenomenon is due to the high probability of sparse domain, that in these way governs the optimization process giving rise to very noisy descent directions. Subsequent techniques are designed to deal with this problem.

%\subsection{\textbf{Only Dense Domain}}
%When training TinyNet just using the dense domain, following this procedure, the resulting model obtains a per-class accuracy of $39.4\%$ 
%in the validation domain (see Table~\ref{tab:training}). Such a result is far from acceptable, since even a classical technique based on 
%hand-crafted features and a CRF~\cite{Valentin13} trained with the same data is able to obtain a $41.3\%$ of per-class accuracy.

\vspace{-3mm}
\subsection{Balanced Gradient Contribution (BGC)}
\label{sec:training_balanced}
The severe statistical difference between the domains induces a large variance in gradients for a sequence of mini-batches. Data from the 
dense domain is more stable and suitable for structural classes, but less informative in general. Data from the sparse domain is highly 
informative, with critical information about dynamic classes, but very noisy. To deal with these aspects we propose to compute search 
directions using the directions proposed by the dense domain under a controlled perturbation given by the sparse domain as shown in 
(\ref{eq:bgc}).

\vspace{-3mm}
\begin{equation}
\text{Loss}_{\text{BGC}}(\mathcal{X}, \mathcal{Y}) = \text{Loss}_\text{WCE}(\mathcal{X}^D, \mathcal{Y}^D) + \lambda 
\text{Loss}_\text{WCE}(\mathcal{X}^S, \mathcal{Y}^S),
\label{eq:bgc}
\end{equation}

\noindent where $\mathcal{X}, \mathcal{Y}$ stand for a subset of samples and their associated labels, drawn from the dense (D) or sparse 
(S) domains. This procedure can be seen as the addition of a very informative regularizer controlled by the parameter $\lambda$, but an 
analogous effect can be achieved by generating mini-batches containing a carefully chosen proportion of images from each domain, such that 
$|\mathcal{X}^D| >> |\mathcal{X}^S|$. This modification of the training procedure leads to superior results and a stable behaviour, as 
reported in section~\ref{sec:exps}.

\subsection{Flying Cars (FC): Domain Adaptation by Data Projection}
\label{sec:training_fc}

Another alternative to solve the problem caused by the combination of incompatible domains is to project or transfer one domain into another. In our case, the noisy sparse domain is projected to the dense domain, using ideas from domain adaptation~\cite{VLP2011b}. This can be achieved, for instance, by selecting random images from the dense domain and using them as backgrounds in which to inject the objects and labels of the sparse domain. This approach was recently used in the ``Flying Chairs" approach of~\cite{Dosovitskiy15} to train DNs for optical flow from synthetic data. It can be seen as a way of performing highly informative data augmentation over the dense domain. Similarly to~\cite{Dosovitskiy15}, we use a naive approach which does not provide a hard constraint on the spatial context of the objects being inserted into the scene, hence the name ``Flying Cars" (FC).

%This transformation of the sparse domain is simple to apply and also ends with the oscillatory problem, but if done naively it tends to break the spatial context (hence the name of Flying-Cars). %In this work we use a simple version of this concept that yet produces good results in practice. 

\vspace{-3mm}
\subsection{Ensemble of Sparse and Dense Domains}\vspace{-1mm}
\label{sec:training_ensemble}
Finally, it is possible to think about the domains as two different tasks: one consisting of recognizing $L_D = 11$ classes from 
finely-annotated data and the other recognizing $L_S = 6$ classes, \ie foreground, traffic signs, poles, cars, pedestrians and cyclists; from noisy sparse annotations. The model trained on the dense domain,  $\theta_D$, is better at structural elements such as roads, 
buildings and sidewalks; while the model trained on the sparse domain, $\theta_S$, is extremely good at segmenting dynamic objects 
such as pedestrians and cyclists. These models can be combined as part of a larger network which adds several new trainable blocks 
to perform a consensus from the output of the original models. In our experiments the ensemble is performed by fixing the original 
networks and adding a convolutional block and four residual-blocks~\cite{He15cvpr} to estimate a consistent output. This is shown in 
Fig.~\ref{fig:cnn_archs}(iv). Residual-blocks were used as they were found to lead to better generalization than simple convolutions in practice. The current configuration, 4 blocks, one with $128$ features and three with $64$ features, was the best configuration we found that did not lead to clear overfitting. This approach is further analysed in section~\ref{sec:exps}.

\vspace{-2mm}

\vspace{-1mm}
\section{Transferring Knowledge across Deconvolutional Networks}
\label{sec:transfer}
\vspace{-1mm}
We use the training methods described in section~\ref{sec:training} to train both FCN and T-Net architectures. The results are reported in 
Table~\ref{tab:training}. For all training methods explored, FCN is observed to consistently outperform the smaller T-Net. %, almost certainly due to its larger capacity.
Moreover, among the different approaches for training, the most outstanding in terms of per-class 
accuracy is the multi-domain ensemble. 
%The reason to prefer per-class accuracy over other statistics is its immunity to class imbalances, a common problem in urban datasets. Thus, we consider it a better model quality indicator.

Despite the high accuracy of the FCN ensemble, its large number of parameters makes it unsuitable for embedded applications, in the context of road scene segmentation. We next investigate whether it is possible to boost a more compact model such as T-Net to have an equivalent performance.
%While T-Net is compact, its accuracy is far from that of the FCN-ensemble. Is this due to an intrinsic capacity limitation of T-Net or some other factor? 
Our hypothesis is that the capacity of T-Net is sufficient to produce results at the level of FCN and FCN-ensemble, but due to specific details of its training and architecture, such as batch normalization and noise within the training data, the methods of section~\ref{sec:training} cannot exploit its full potential. %Some evidence of this can be drawn from Table~\ref{tab:training}, as T-Net BGC outperforms T-Net e2e. 
We therefore examine an alternative training approach for T-Net. We adopt the FCN ensemble as a Source Network (S-Net) and attempt to emulate its behaviour with (\ie transfer its knowledge to) the T-Net. We describe three approaches to transfer knowledge: (i) via labels (TK-L), (ii) via soft-max probabilities (TK-SMP), %and a variation of the previous, 
and (iii) via soft-max probabilities with weighted-cross-entropy (TK-SMP-WCE).

%A better way to exploit the capacity of T-Net is to train it by transferring knowledge from another network, such as the FCN-ensemble, from now on referred to as the Source Network (S-Net). The S-Net has already distilled the discrepancies of the dataset into a more convenient and stable representation. We therefore explore different methods to exploit the ``stable knowledge'' of the S-Net to train a T-Net. These methods are (i) transferring knowledge through labels (TK-L), (ii) transferring knowledge through soft-max probabilities (TK-SMP),%and a variation of the previous, and (iii) transferring knowledge through soft-max probabilities with weighted-cross-entropy (TK-SMP-WCE).

\paragraph{\bf Transferring Knowledge Through Labels (TK-L).}\label{sec:transfer_tktl}
\vspace{-2mm}
This strategy aims to distill the knowledge of the S-Net directly from its predicted labels, in the spirit of~\cite{Bucilua06kdd}. We use both dense and sparse domains of training data described in section~\ref{sec:datasets}, ignoring their original annotations. %Instead, the smooth labels predicted by the S-Net serve to train the T-Net. 
The benefit of this approach is that the multi-modality of the data has been filtered by the S-Net and some distractors are ignored, so the information reaching T-Net is simpler, leading to a smoother search space and making it easier to find good solutions. In our setup we included extra training data from a large unlabelled Google Street View (GSV) dataset~\cite{Zamir14pami}, taken of street scenes from multiple cities in the US. We remove the upward facing camera and took a random crop from each image to produce 51,715 images. We combined previous and new training data using BGC to train the T-Net with a standard cross-entropy loss. Here, BGC is used as an important mechanism to control the influence of the GSV data and prevent from drift.

\paragraph{\bf Transferring Knowledge Through SoftMax Probabilities (TK-SMP).}\label{sec:transfer_tktsp}
\vspace{-2mm}
The strategy uses additional information from S-Net during transfer, by considering the probability distributions produced by the softmax classifier, which contains information about how different classes are correlated~\cite{HintonVinyals15}. To this end, we train a T-Net using standard cross-entropy between the probability distributions of S-Net and T-Net as our loss function. As in the previous strategy, the training makes use of BGC to control the influence of GSV data to bound its contribution. This second approach leads to a notable improvement of the network per-class accuracy as shown in Table~\ref{tab:transfer} (i)-(ii). 

A variation of this method consists of adding drop-out blocks to the T-Net during the transference process. In practice, this addition behaves as in end-to-end training, helping to improve the generalization of the net. See Table~\ref{tab:transfer}(iii).

\paragraph{\bf Transferring Knowledge Through SoftMax Probabilities with WCE (TK-SMP-WCE).}\label{sec:transfer_tktspwce}
\vspace{-2mm}
One of the problems with the previous approaches of TK-TL and TK-SMP is that they do not account for class imbalance during transfer. In practice this means that the resulting models are biased towards the dominant classes and producing models with higher per-class accuracy requires a higher number of epochs during training. We propose to solve this problem by controlling the influence of each softmax sample with WCE, in the same way that the influence of different datasets is controlled by BGC. This simple modification, in combination with the use of dropout in the T-Net, leads to models that have virtually the same per-class accuracy as the S-Net, \ie an ensemble of FCNs; see Table~\ref{tab:transfer} (iv). In this way the full potential of the T-Net is unlocked, giving rise to an accurate and memory-efficient model, convenient for embedded systems and automotive applications.

% Experimental Results
\section{Experimental Results}\label{sec:exps}
\setlength{\medmuskip}{0mu}
\vspace{-2mm}

We evaluate the performance of the proposed training methodology with respect to a set of state-of-the-art baselines. 
Special emphasis is set on the performance of our TK-SMP-WCE transfer technique when used in combination with Balanced Gradient 
Contributions (BGC).

\vspace{-2mm}\paragraph{\bf Experiment Setup.}
All our experiments are carried out on the MDRS3 dataset (Section~\ref{sec:datasets}), testing on the combination of U-LabelMe 
and CBCL (1,526 images overall). Due to time and resource constraints, we subsample the original images to a resolution of 
$240\times180$ in all our experiments. This speeds up training and evaluation of models but makes certain classes, such as 
sidewalks, poles and traffic signs, systematically harder to recognize for all models due to the low resolution. Nevertheless, 
this factor is consistent across all the experiments and does not affect the conclusions obtained when comparing different training 
approaches and models. Images are initially normalized using spatial contrast normalization, independently applied to each 
channel. Afterwards, zero-mean and data re-scaling in the range [-127,127] are applied. In practice we observed that this 
normalization speeds-up convergence.

Results are evaluated according to the average per-class accuracy (\textbf{per-class}) and the global accuracy (\textbf{global}). Given the number of pixels, $n_{i,j}$, belonging to class $i$ and classified as class $j$, and assuming $L$ is the number of classes, then \textbf{per-class} is evaluated as $\frac{1}{L}\sum_i n_{i,i}/\sum_{j=1}^{L} n_{i,j}$ and \textbf{global} as %$\sum_i n_{i,i}/\sum_{i}^{K} \sum_{j=1}^{K} n_{i,j}$
$\sum_i n_{i,i}/N$ where $N$ is the total number of pixels in the evaluation set. Due to the intrinsic unbalanced nature of the class frequencies in urban scenes, 
we consider the average per-class to be more significant to assess the recognition and generalisation capabilities of the models. Within parenthesis we report the difference between the results of the current method and the FCN model at Table~\ref{tab:training}(i) as a reference (improvements are highlighted in blue, diminishments in red).

%All our experiments are evaluated over the datasets of U-LabelMe and CBCL consisting of a total of 4489 challenging images of different cities (\eg Barcelona, Madrid, Boston, etc.). The training data is composed by two datasets. The first source consists of finely pixel-wise annotated data from the CamVid and the KITTI semantics datasets. The second source aggregates data coming from MS COCO, KITTI-O, ETH-RMPTMP and GTSRB datasets, consisting of sparsely annotated data containing a couple of foreground objects per image. Additionally, the unlabelled data of the Google Street View Data Set~\cite{GMCP} is used when knowledge transfer is applied. Due to time and resources constraints, we subsample the original images to a resolution of $240\times 180$ in all our experiments. This speedsup training and evaluation of models but makes certain classes, such as sidewalks, poles and traffic signs, systematically harder to recognize for all models due to the low resolution. Nevertheless, this factor is consistent across all the experiments and does not affect the conclusions obtained when compared different training approaches and models.

\vspace{-2mm}
\subsection{Assessing Multi-Domain Training}\label{sec:exps:1}
%\vspace{-2mm}

\setlength{\textfloatsep}{10pt plus 10.0pt minus 20.0pt}
\setlength{\floatsep}{10pt plus 10.0pt minus 20.0pt}
\setlength{\intextsep}{12.0pt plus 20.0pt minus 20.0pt}
%\vspace{-5mm}

\paragraph{\bf End-to-End training.}

In Table~\ref{tab:training} (i)-(iii), we first evaluate the performance of T-Net against the FCN network~\cite{Long15cvpr} and ALE~\cite{Ladicky10eccv}, a classical semantic segmentation framework based on hand-crafted features. These models are trained using the dense domain 
only, with the end-to-end approach described in section~\ref{sec:training_endtoend}. As Table~\ref{tab:training} (i)-(iii) shows, for this initial setup T-Net underperforms both FCN, by 11.2 points per-class, and ALE, by 1.9 points.%  and it is even worse than ALE, a result that may be wrongly attributed to T-Net  low capacity.

We extended this first evaluation by adding the sparse domain to the end-to-end training. However, as shown in Table~\ref{tab:training} (iv) and (viii) the 
training diverged in both cases. This phenomenon was commented on in section~\ref{sec:training_endtoend} and is attributed to the gradient noise 
introduced by the sparse domain when its contribution is unbounded. This reinforces our claim that control over the distribution and the complexity of the data is required to produce competitive training results. 

\paragraph{\bf Flying Cars, BGC \& Ensemble.}
 When the end-to-end training is replaced by methods implementing policies to control the contribution of each domain, the improvement in accuracy is notable. Table~\ref{tab:training} (v)-(vii) shows that for all the techniques,  controlled training improves the per-class of the standard FCN. FC and BGC methods, although not achieving the top performance, have the advantage of requiring just one training stage; while the ensemble requires training individual models first (per domain) and then merging them. Yet, since the ensemble of FCN achieves the highest performance we use it as our S-Net, and try to match its performance with T-Net. The outcome of applying FC, BGC and ensemble on the T-Net are analogous to the previous case; and again, the ensemble renders the best results in terms of per-class accuracy (see Table~\ref{tab:training} (ix)-(xi)).

%%%%%%%%%%%%%%%%%%%%%%%%%%%%%%%%%%%%%%%%%%%%%%%%%%%%%%%%%%%%%%%%%%%%%%%%%%%%%%%%%%%%%%%%%
\subsection{Evaluation of Knowledge Transfer Methods}\label{sec:exps:3}
%%%%%%%%%%%%%%%%%%%%%%%%%%%%%%%%%%%%%%%%%%%%%%%%%%%%%%%%%%%%%%%%%%%%%%%%%%%%%%%%%%%%%%%%%
As summarized in Table~\ref{tab:transfer}, results of previous training approaches on T-Net are dramatically improved when applying knowledge 
transfer methods. For all the transference method we added the unlabelled data from the Google Street View Data Set~\cite{GMCP} in order to 
increase the variability of the S-Net responses during the process, which helps capturing the behaviour of S-Net. 

Here we see that the evolution of the transferring techniques is directly correlated to the improvement of the T-Net performance. A simple 
transfer of labels (TK-L) from the S-Net produces a T-Net model that is already 2.9 points better than FCN (used here as a reference). When 
the transfer is based on the softmax probability distribution over the classes, as in TK-SMP, accuracy is boosted up to 57.3 (6.7 points 
better than FCN). It is worth noticing than, when dropout is included in the TK-SMP transference (TK-SMP-Drop), it improves global accuracy 
in 3.2 points compared to FCN. We observed this effect when using dropout at expense of some loss in per-class accuracy. 

\begin{figure}[t!]
	\centering
	\includegraphics[width=0.95\textwidth]{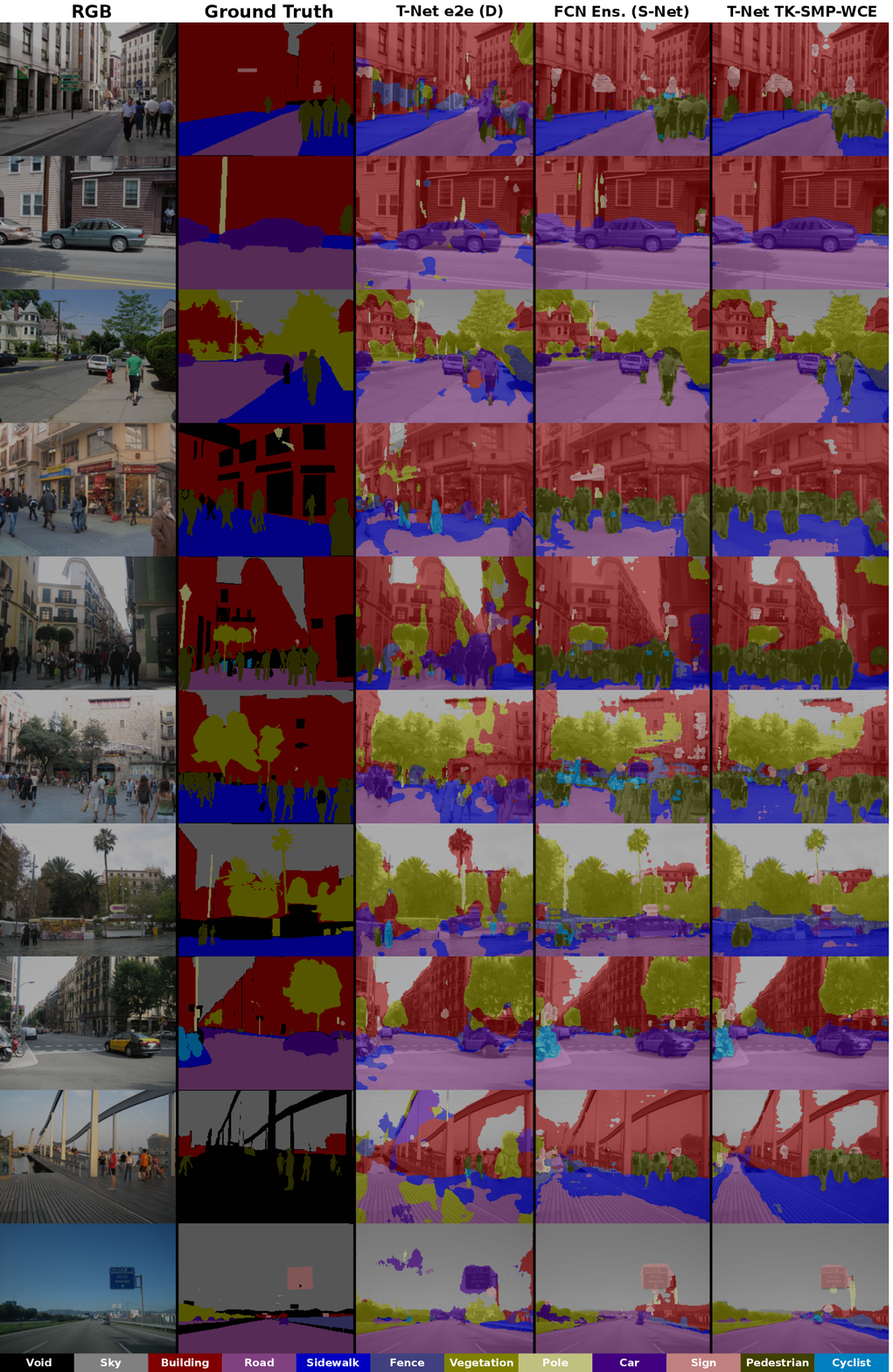}
	\vspace{-4mm}
	\caption{Qualitative results on test images for different training methods. Our proposed method of training T-Net via transfer learning results in visually good segmentations, in some cases providing better results than the FCN ensemble and even noisy ground truth.}
	\label{fig:examples}
\end{figure}

Finally, Table~\ref{tab:transfer} (iv) shows that when the S-Net softmax distributions are weighted according to their 
relevance in the dataset (\ie less abundant more relevant), the transference of this knowledge reaches the maximum performance found so far, 
59.3\% of per-class accuracy. Thus, the TK-SMP-WCE approach produces a T-Net 9.1 points better than FCN in per-class and 0.2 in global 
accuracy, almost reaching the results of the S-Net, \ie an ensemble of two FCN which has $200\times$ more parameters. Visual results of this evolution 
are shown in Fig.~\ref{fig:examples} for testing examples. Notice how the T-Net obtained from TK-SMP-WCE can sometimes produce better 
results than the S-Net. We believe that these results give strong evidence to render knowledge transfer methods and in particular TK-SMP-WCE 
as preferred methods to train memory constrained DNs. 

\begin{table}[t!]
	\vspace{2mm}
	\centering
	\caption{Semantic segmentation quantitative results for FCN and T-Net on the testing dataset for the training methods under study, \ie end-to-end (dense, D; sparse, S; and both, D+S), BGC, Flying cars (FC) and net ensemble (Ens.).}
	\vspace{-3mm}
	\resizebox{\textwidth}{!}{%
		\scriptsize
		\tabcolsep=0.06cm
		\begin{tabular}{@{}lcccccccccccll@{}}
			\toprule
			& \includegraphics{figs/sky.eps} & \includegraphics{figs/building.eps} & \includegraphics{figs/road.eps} & \includegraphics{figs/sidewalk.eps} & \includegraphics{figs/fence.eps} & \includegraphics{figs/trees.eps} & \includegraphics{figs/pole.eps} & \includegraphics{figs/car.eps} & \includegraphics{figs/traffic.eps} & \includegraphics{figs/pedestrian.eps} & \includegraphics{figs/cyclist.eps} & \\
			Method    & sky & building & road & sidewalk & fence & vegetat. & pole & car & sign & pedest. & cyclist & \textbf{per-class} & \textbf{global} \\ \midrule
			(i) FCN~\cite{Long15cvpr} (D)					& 77.8    & 67.6     & 86.1   & 35.9      & 35.1     & 89.6    & 8.9    & 86.2   & 41.3    & 17.6         & 10.6      	& 50.6              & 71.6              \\ \midrule
			(ii) T-Net (D)  									& 80.6    & 62.0     & 85.2   & 20.6      & 4.4     & 84.5         & 9.4    & 70.0  & 6.4   & 7.6         & 2.3       & 39.4 \UP{-11.2}       & 66.6 \UP{-5.0}         \\ \midrule
			
			(iii) ALE~\cite{Ladicky10eccv} (D) 				& 85.0  	& 69.0      & 94.0   & 8.0       & 19.0    & 83.0         & 3.0    & 74.0  & 13.0    & 5.0        & 1.0       & 41.3 \UP{-9.3}              & 72.1   \UP{0.5}   \\ 
			\midrule\midrule
			(iv) FCN~\cite{Long15cvpr} e2e (D+S)  	& \multicolumn{13}{c}{\textbf{training diverged}}\\ \midrule
			(v) FCN~\cite{Long15cvpr} FC 	  				& 85.3   	& 71.4     & 87.0   & 26.4      & 19.8     & 86.5    & 10.6   & 89.3   & 45.4    & 58.8         & 7.0       & 53.4  \UP{2.8}                & 73.2 \UP{1.6}              \\ \midrule
			(vi) FCN~\cite{Long15cvpr} BGC 	  				& 80.3   	& 73.5     & 82.6   & 49.5      & 39.6     & 91.6    & 11.6    & 87.3   & 50.8    & 44.1         & 19.6      & 57.3  \UP{6.7}                & 75.5   \UP{3.9}            \\ \midrule
			(vii) FCN~\cite{Long15cvpr} Ens. (\textbf{S-Net}) 	& 77.4   	& 71.9     & 85.0   & 27.8      & 40.8     & 85.8    & 8.0    & 93.4   & 43.0    & 80.4         & 60.6      & 61.3   \UP{10.7}               & 73.4    \UP{1.8}           \\ \midrule \midrule
			(viii) T-Net e2e (D+S) 	  & \multicolumn{13}{c}{\textbf{training diverged}}\\ \midrule
			(ix) T-Net FC   									& 77.5    & 67.2     & 77.7   & 34.4      & 18.4     & 86.3         & 8.0    & 80.0  & 18.8   & 25.9         & 4.8       & 45.3  \UP{-5.3}             & 69.1	\UP{-2.5}			\\ \midrule
			(x) T-Net BGC   									& 58.9    & 64.5     & 81.6   & 21.5      & 4.8     & 83.1         & 11.0    & 82.3  & 21.2   & 31.3         & 9.3       & 42.7   \UP{-8.3}            & 64.9		\UP{-6.7}		\\ \midrule
			(xi) T-Net Ens.   								& 89.0    & 57.4     & 85.5   & 22.9      & 0.3     & 92.2         & 11.4    & 86.3  & 14.6   & 56.6         & 16.9      & 46.9   \UP{-3.7}            & 65.5	\UP{-6.1}			\\ \midrule
			
		\end{tabular}%
	}
	\label{tab:training}
\end{table}

%\setlength{\textfloatsep}{10pt plus 1.0pt minus 12.0pt}
%\setlength{\floatsep}{10pt plus 1.0pt minus 12.0pt}
%\setlength{\intextsep}{12.0pt plus 2.0pt minus 8.0pt}
%\vspace{-8mm}

% Experiment 3
\begin{table*}[t!]
\vspace{-2mm}
	\centering
	\caption{Evaluation of the proposed Knowledge Transfer techniques for $\text{S-Net} \rightarrow \text{T-Net}$.}
	\vspace{-3mm}
	\label{tab:exp:3}
	\resizebox{\textwidth}{!}{%
		\scriptsize
		\tabcolsep=0.06cm
		\begin{tabular}{@{}lccccccccccccc@{}}
			\toprule
			& \includegraphics{figs/sky.eps} & \includegraphics{figs/building.eps} & \includegraphics{figs/road.eps} & \includegraphics{figs/sidewalk.eps} & \includegraphics{figs/fence.eps} & \includegraphics{figs/trees.eps} & \includegraphics{figs/pole.eps} & \includegraphics{figs/car.eps} & \includegraphics{figs/traffic.eps} & \includegraphics{figs/pedestrian.eps} & \includegraphics{figs/cyclist.eps} & \\
			Method    & sky & building & road & sidewalk & fence & vegetat. & pole & car & sign & pedest. & cyclist & \textbf{per-class} & \textbf{global} \\ \midrule
			
			%FCN-Ens $\rightarrow$ FCN* 	  			& 73.0   	& 73.4     & 74.4   & 23.9      & 12.9     & 85.6    & 3.4    & 91.8   & 40.6    & 73.0         & 57.2      & 60.6                  & 69.8               \\ \midrule
			% S-Net $\rightarrow$ removed
		(b) baseline: FCN~\cite{Long15cvpr} (D)					& 77.8    & 67.6     & 86.1   & 35.9      & 35.1     & 89.6    & 8.9    & 86.2   & 41.3    & 17.6         & 10.6      	& 50.6              & 71.6              \\ \midrule
			(i) T-Net (TK-L)    & 88.7   	& 50.6     & 68.8   & 45.4      & 48.7     & 77.2    & 18.6    & 73.1   & 19.4    & 68.8         & 29.3      & 53.5 \UP{2.9}                 & 62.9 \UP{-8.7}              \\ \midrule
			(ii) T-Net (TK-SMP)       & 85.1   	& 65.1     & 87.5   & 21.1      & 35.7     & 85.3    & 6.6   & 90.0   & 45.2    & 53.2         & 55.6     & 57.3  \UP{6.7}            & 70.8    \UP{-0.8}           \\ \midrule
			%T-Net $\rightarrow$ P-Net (TK-SMP-Drop)       & 81.8   	& 65.2     & 74.9   & 34.1      & 3.6     & 94.1    & 2.0    & 91.6   & 33.5    & 63.1         & 73.1      & 57.1                  & 70.9               \\ \midrule
			(iii) T-Net (TK-SMP-Drop)   & 87.6   	& 75.9     & 79.3   & 43.2      & 27.1     & 80.8    & 4.0    & 86.9   & 19.9    & 68.5         & 14.0      & 53.4  \UP{2.8}                & 74.8  \UP{3.2}             \\ \midrule
			%T-Net $\rightarrow$ P-Net (Probs-drop2)   & 85.8   	& 63.8     & 86.4   & 25.5      & 21.9     & 88.9    & 2.3    & 83.3   & 18.6    & 75.5         & 50.0      & 54.7                  & 70.7               \\ \midrule
			(iv) T-Net (TK-SMP-WCE)   & 87.4 & 66.9 & 82.0 & 33.0 & 37.9 & 83.3 & 14.1 & 89.4 & 40.0 & 78.6 & 40.2 & 59.3 \UP{9.1} & 71.8  \UP{0.2}            \\ \midrule
			
		\end{tabular}%
	}
	\label{tab:transfer}
	%\vspace{-4mm}
\end{table*}

%\setlength{\textfloatsep}{10pt plus 1.0pt minus 16.0pt}
%\setlength{\floatsep}{10pt plus 1.0pt minus 16.0pt}
%\setlength{\intextsep}{12.0pt plus 2.0pt minus 12.0pt}
%\vspace{-2mm}

%\vspace{-3mm}
% Conclusions and Future Work
%******************************************************************
\vspace{7mm}
\section{Conclusions and Future Work}\label{sec:conclusions}\vspace{-2mm}
%\vspace{-2mm}
In this work we have described a training strategy for DNs to be used in resource-constrained applications such as road scene segmentation. We showed that training relatively shallow target networks via regular end-to-end approaches on a challenging aggregate dataset leads to underperformance versus state-of-the-art deep models. One likely cause for this is that shallow models are much harder to optimize and, when trained directly on noisy or multi-modal data, have difficulty in navigating local minima.
To overcome this, we extended the idea of knowledge transfer to DNs. We first explored various means of producing a best-performing source network, by relaxing resource constraints and ensembling networks across different data domains.
We then demonstrated that by using the source network as a guide, it was possible to train a target network which satisfied all constraints while giving better performance than a state-of-the-art FCN network and almost the same performance than an ensemble of FCNs, with a memory footprint that is just $0.5\%$ of the ensemble.
%Why does this give better results? The source network provides softer labels which may be easier to learn than hard ground truth labels. Also possibly cleaner labels (if there are errors in the ground truth).
We believe that our findings will be very useful for training DNs not just in automotive settings but also in context where labelled data is limited and practical constraints exist on model size.

%In this work we proposed a "recipe book", or set of steps to improve accuracy, compactness and computational performance of DeconvNets models for semantic segmentation in Automotive Environments. We showed that, given the current constraints on the amount and type of the available training data and on the memory limitations of embedded devices, performing standard end-to-end training leads to dubious results. This situation is addressed in our "recipe book" by (i) generation and gathering of sources of data from object detection and semantic segmentation domains; (ii) training individual FCNs on the different domains and then combine them via ensemble (Teacher-Net); and (iii) transfer the knowledge from the unconstrained Teacher-Net to a time-and-memory efficient pooling-unpooling TinyNet, that is 100 times smaller. This final network meets the desired memory and time constraints present in AE environments while keeping the high accuracy of the Teacher-Net. We believe that the findings contained in this "recipe book" can be very useful for \textit{AE} scenarios and also in those settings where labelled data is diverse and limited, and where practical constraints exist on model size.

%*******************************************************************************
\clearpage

\bibliographystyle{splncs}
\bibliography{references}

\begin{thebibliography}{10}

\bibitem{Farabet13pami}
Farabet, C., Couprie, C., Najman, L., LeCun, Y.:
\newblock Learning hierarchical features for scene labeling.
\newblock {IEEE} Trans. Pattern Anal. Machine Intell. \textbf{35}(8) (2013)
  1915--1929

\bibitem{Long15cvpr}
Long, J., Shelhamer, E., Darrell, T.:
\newblock Fully convolutional networks for semantic segmentation.
\newblock In: IEEE Conf. on Computer Vision and Pattern Recognition (CVPR).
  (2015)

\bibitem{Noh15iccv}
Noh, H., Hong, S., Han, B.:
\newblock Learning deconvolution network for semantic segmentation.
\newblock In: Intl. Conf. on Computer Vision (ICCV). (2015)

\bibitem{Badrinarayanan15arxiv}
Badrinarayanan, V., Kendall, A., Cipolla, R.:
\newblock {SegNet}: A deep convolutional encoder-decoder architecture for image
  segmentation.
\newblock arXiv preprint arXiv:1511.00561 (2015)

\bibitem{Lin14eccv}
Lin, T.Y., Maire, M., Belongie, S., Hays, J., Perona, P., Ramanan, D.,
  Doll\'{a}r, P., Zitnick, C.L.:
\newblock {Microsoft COCO: Common Objects in Context}.
\newblock In: Eur. Conf. on Computer Vision (ECCV). (2014)

\bibitem{Russakovsky15ijcv}
Russakovsky, O., Deng, J., Su, H., Krause, J., Satheesh, S., Ma, S., Huang, Z.,
  Karpathy, A., Khosla, A., Bernstein, M., Berg, A.C., Fei-Fei, L.:
\newblock {ImageNet} large scale visual recognition challenge.
\newblock Intl. J. of Computer Vision (2015)

\bibitem{MITPLACES}
Zhou, B., Lapedriza, A., Xiao, J., Torralba, A., Oliva, A.:
\newblock Learning deep features for scene recognition using places database.
\newblock In: Advances in Neural Information Processing Systems (NIPS).
\newblock (2014)

\bibitem{Brostow09prl}
Brostow, G.J., Fauqueur, J., Cipolla, R.:
\newblock Semantic object classes in video: A high-definition ground truth
  database.
\newblock Pattern Recognition Letters \textbf{30}(2) (2009)  88--97

\bibitem{Geiger13ijrr}
Geiger, A., Lenz, P., Stiller, C., Urtasun, R.:
\newblock Vision meets robotics: The {KITTI} dataset.
\newblock Intl. J. of Robotics Research (2013)

\bibitem{Chen15cvpr}
Chen, Q., Huang, J., Feris, R., Brown, L.M., Dong, J., Yan, S.:
\newblock Deep domain adaptation for describing people based on fine-grained
  clothing attributes.
\newblock In: IEEE Conf. on Computer Vision and Pattern Recognition (CVPR).
  (2015)

\bibitem{Tzeng15iccv}
Tzeng, E., Hoffman, J., Darrell, T., Saenko, K.:
\newblock Simultaneous deep transfer across domains and tasks.
\newblock In: Intl. Conf. on Computer Vision (ICCV). (2015)

\bibitem{HintonVinyals15}
Hinton, G., Vinyals, O., Dean, J.:
\newblock Distilling the knowledge in a neural network.
\newblock arXiv preprint arXiv:1503.02531 (2015)

\bibitem{Kundu14eccv}
Kundu, A., Li, Y., Dellaert, F., Li, F., Rehg, J.M.:
\newblock Joint semantic segmentation and {3D} reconstruction from monocular
  video.
\newblock In: Eur. Conf. on Computer Vision (ECCV). (2014)

\bibitem{Ros15wacv}
Ros, G., Ramos, S., Granados, M., Bakhtiary, A., V\'{a}zquez, D., L\'{o}pez,
  A.M.:
\newblock Vision-based offline-online perception paradigm for autonomous
  driving.
\newblock In: {Winter Conference on Applications of Computer Vision (WACV)}.
  (2015)

\bibitem{Silberman12}
N.~Silberman, D.H., Kohli, P., Fergus, R.:
\newblock Indoor segmentation and support inference from {RGBD} images.
\newblock In: Eur. Conf. on Computer Vision (ECCV). (2012)

\bibitem{Handa16cvpr}
Handa, A., Patraucean, V., Badrinarayanan, V., Stent, S., Cipolla, R.:
\newblock Understanding real world indoor scenes with synthetic data.
\newblock In: IEEE Conf. on Computer Vision and Pattern Recognition (CVPR).
  (2016)

\bibitem{Tighe10eccv}
Tighe, J., Lazebnik, S.:
\newblock Superparsing: Scalable nonparametric image parsing with superpixels.
\newblock In: Eur. Conf. on Computer Vision (ECCV). (2010)

\bibitem{Sengupta13icra}
Sengupta, S., Greveson, E., Shahrokni, A., Torr, P.H.S.:
\newblock Urban {3D} semantic modelling using stereo vision.
\newblock In: IEEE Intl. Conf. on Robotics and Automation (ICRA). (2013)

\bibitem{Hu13icra}
Hu, H., Munoz, D., Bagnell, J.A., Hebert, M.:
\newblock Efficient 3-{D} scene analysis from streaming data.
\newblock In: IEEE Intl. Conf. on Robotics and Automation (ICRA). (2013)
  2297--2304

\bibitem{Kohli09ijcv}
Kohli, P., Ladick\'{y}, L., Torr, P.H.S.:
\newblock Robust higher order potentials for enforcing label consistency.
\newblock Intl. J. of Computer Vision \textbf{82}(3) (2009)  302--324

\bibitem{Ladicky10eccv}
Ladick\'{y}, L., Sturgess, P., Alahari, K., Russell, C., Torr, P.H.S.:
\newblock What, where and how many? {C}ombining object detectors and {CRF}s.
\newblock In: Eur. Conf. on Computer Vision (ECCV). (2010)  427--437

\bibitem{Valentin13}
Valentin, J.P.C., Sengupta, S., Warrell, J., Shahrokni, A., Torr, P.H.S.:
\newblock Mesh based semantic modelling for indoor and outdoor scenes.
\newblock In: IEEE Conf. on Computer Vision and Pattern Recognition (CVPR).
  (2013)

\bibitem{Alvarez12eccv}
\'{A}lvarez, J.M., Gevers, T., LeCun, Y., L\'{o}pez, A.M.:
\newblock Road scene segmentation from a single image.
\newblock In: Eur. Conf. on Computer Vision (ECCV). (2012)

\bibitem{Girshick14rcnn}
Girshick, R., Donahue, J., Darrell, T., Malik, J.:
\newblock Rich feature hierarchies for accurate object detection and semantic
  segmentation.
\newblock In: IEEE Conf. on Computer Vision and Pattern Recognition (CVPR).
  (2014)

\bibitem{icml2015_chenb15}
Chen, L.C., Schwing, A., Yuille, A., Urtasun, R.:
\newblock Learning deep structured models.
\newblock In: Intl. Conf. on Machine Learning (ICML). (2015)

\bibitem{Zheng15iccv}
Zheng, S., Jayasumana, S., Romera-Paredes, B., Vineet, V., Su, Z., Du, D.,
  Huang, C., Torr, P.H.S.:
\newblock Conditional random fields as recurrent neural networks.
\newblock In: Intl. Conf. on Computer Vision (ICCV). (2015)

\bibitem{Dai15iccv}
Dai, J., He, K., Sun, J.:
\newblock {BoxSup}: Exploiting bounding boxes to supervise convolutional
  networks for semantic segmentation.
\newblock In: Intl. Conf. on Computer Vision (ICCV). (2015)

\bibitem{Papandreu15iccv}
Papandreou, G., Chen, L.C., Murphy, K., Yuille, A.L.:
\newblock Weakly- and semi-supervised learning of a deep convolutional network
  for semantic image segmentation.
\newblock In: Intl. Conf. on Computer Vision (ICCV). (2015)

\bibitem{Everingham15}
Everingham, M., Eslami, S.M.A., Van~Gool, L., Williams, C.K.I., Winn, J.,
  Zisserman, A.:
\newblock The pascal visual object classes challenge: A retrospective.
\newblock Intl. J. of Computer Vision \textbf{111} (2015)

\bibitem{Cordts2015Cvprw}
Cordts, M., Omran, M., Ramos, S., Scharw{\"a}chter, T., Enzweiler, M.,
  Benenson, R., Franke, U., Roth, S., Schiele, B.:
\newblock The {C}ityscapes dataset.
\newblock In: CVPR Workshop on The Future of Datasets in Vision. (2015)

\bibitem{Gros2016}
Ros, G., Sellart, L., Materzynska, J., Vazquez, D., Lopez, A.:
\newblock {SYNTHIA}: A large collection of synthetic images for semantic
  segmentation of urban scenes.
\newblock In: IEEE Conf. on Computer Vision and Pattern Recognition (CVPR).
  (2016)

\bibitem{He15cvpr}
He, K., Zhang, X., Ren, S., Sun, J.:
\newblock Deep residual learning for image recognition.
\newblock In: IEEE Conf. on Computer Vision and Pattern Recognition (CVPR).
  (2016)

\bibitem{Ba14nips}
Ba, L.J., Caruana, R.:
\newblock Do deep nets really need to be deep?
\newblock In: Advances in Neural Information Processing Systems (NIPS). (2014)

\bibitem{Choromanska15aistats}
Choromanska, A., Henaff, M., Mathieu, M., Ben~Arous, G., LeCun, Y.:
\newblock The loss surfaces of multilayer networks.
\newblock In: Proceedings of the Eighteenth International Conference on
  Artificial Intelligence and Statistics. (2015)  192--204

\bibitem{Bucilua06kdd}
Bucilu\u{a}, C., Caruana, R., Niculescu-Mizil, A.:
\newblock Model compression.
\newblock In: Proceedings of the 12th ACM SIGKDD international conference on
  Knowledge discovery and data mining, ACM (2006)  535--541

\bibitem{Romero15-iclr}
Romero, A., Ballas, N., Kahou, S.E., Chassang, A., Gatta, C., Bengio, Y.:
\newblock Fitnets: Hints for thin deep nets.
\newblock In: Intl. Conf. on Learning Representations (ICLR). (2015)

\bibitem{HanMD15}
Han, S., Mao, H., Dally, W.J.:
\newblock Deep compression: Compressing deep neural network with pruning,
  trained quantization and huffman coding.
\newblock CoRR (2015)

\bibitem{Iandola16arxiv}
Iandola, F.N., Moskewicz, M.W., Ashraf, K., Han, S., Dally, W.J., Keutzer, K.:
\newblock {SqueezeNet: AlexNet-level accuracy with 50$\times$ fewer parameters
  and $<$1MB model size}.
\newblock arXiv preprint arXiv:1602.07360 (2016)

\bibitem{Brostow08eccv}
Brostow, G.J., Shotton, J., Cipolla, R.:
\newblock Segmentation and recognition using structure from motion point
  clouds.
\newblock In: Eur. Conf. on Computer Vision (ECCV). (2008)

\bibitem{Russell08ijcv}
Russell, B.C., Torralba, A., Murphy, K.P., Freeman, W.T.:
\newblock {LabelMe}: a database and web-based tool for image annotation.
\newblock Intl. J. of Computer Vision \textbf{77}(1--3) (2008)  157--173

\bibitem{Bileschi07cbcl}
Bileschi, S.:
\newblock {CBCL StreetScenes} challenge framework.
\newblock \url{http://cbcl.mit.edu/software-datasets/streetscenes/} (2007)

\bibitem{Ess09pami}
Ess, A., Leibe, B., Schindler, K., Gool, L.V.:
\newblock Robust multi-person tracking from a mobile platform.
\newblock {IEEE} Trans. Pattern Anal. Machine Intell. \textbf{31}(10) (2009)
  1831--1846

\bibitem{Houben13ijcnn}
Houben, S., Stallkamp, J., Salmen, J., Schlipsing, M., Igel, C.:
\newblock Detection of traffic signs in real-world images: The {G}erman
  {T}raffic {S}ign {D}etection {B}enchmark.
\newblock In: {International Joint Conference on Neural Networks}. Number 1288
  (2013)

\bibitem{Simonyan15iclr}
Simonyan, K., Zisserman, A.:
\newblock Very deep convolutional networks for large-scale image recognition.
\newblock In: Intl. Conf. on Learning Representations (ICLR). (2015)

\bibitem{Zeiler10cvpr}
Zeiler, M.D., Krishnan, D., Taylor, G.W., Fergus, R.:
\newblock Deconvolutional networks.
\newblock In: IEEE Conf. on Computer Vision and Pattern Recognition (CVPR).
  (2010)

\bibitem{Ioffe15icml}
Ioffe, S., Szegedy, C.:
\newblock Batch normalization: Accelerating deep network training by reducing
  internal covariate shift.
\newblock In: Intl. Conf. on Machine Learning (ICML). (2015)

\bibitem{Chatfield14bmvc}
Chatfield, K., Simonyan, K., Vedaldi, A., Zisserman, A.:
\newblock Return of the devil in the details: Delving deep into convolutional
  networks.
\newblock In: British Machine Vision Conf. (BMVC). (2014)

\bibitem{He15iccv}
He, K., Zhang, X., Ren, S., Sun, J.:
\newblock Delving deep into rectifiers: Surpassing human-level performance on
  {ImageNet} classification.
\newblock In: Intl. Conf. on Computer Vision (ICCV). (2015)

\bibitem{LeNCLPN11}
Le, Q.V., Ngiam, J., Coates, A., Lahiri, A., Prochnow, B., Ng, A.Y.:
\newblock On optimization methods for deep learning.
\newblock In: Intl. Conf. on Machine Learning (ICML). (2011)

\bibitem{VLP2011b}
Vazquez, D., Lopez, A., Ponsa, D., Marin, J.:
\newblock Cool world: domain adaptation of virtual and real worlds for human
  detection using active learning.
\newblock In: NIPS Domain Adaptation Workshop: Theory and Application. (2011)

\bibitem{Dosovitskiy15}
Dosovitskiy, A., Fischer, P., Ilg, E., Hausser, P., Hazirbaş, C., Golkov, V.,
  Smagt, P.V., Cremers, D., Brox, T.:
\newblock Flownet: Learning optical flow with convolutional networks.
\newblock In: Intl. Conf. on Computer Vision (ICCV). (2015)

\bibitem{Zamir14pami}
Zamir, A.R., Shah, M.:
\newblock Image geo-localization based on multiplenearest neighbor feature
  matching usinggeneralized graphs.
\newblock Pattern Analysis and Machine Intelligence, IEEE Transactions on
  \textbf{36}(8) (2014)  1546--1558

\bibitem{GMCP}
Zamir, A., Shah, M.:
\newblock Image geo-localization based on multiple nearest neighbor feature
  matching using generalized graphs.
\newblock (2014)

\end{thebibliography}

%\clearpage
%\input{srcs/supmat.tex}

\end{document}